\documentclass[10pt,twocolumn,twoside] {IEEEtran}
\usepackage{graphicx,amsfonts,amsmath,amssymb,xcolor,amsthm}
\usepackage{algorithm,algorithmic,bm,color}
\usepackage{multirow}
\usepackage{arydshln}
\usepackage[noadjust]{cite}
 \def\comment#1{}
\newcommand{\argmin}{\arg\!\min}

\newcommand{\Dmat}{{\bf D}}

\newcommand{\Hmat}[0]{{{\bf H}}}

\newcommand{\Rmat}[0]{{{\bf R}}}
\newcommand{\Smat}[0]{{{\bf S}}}

\newcommand{\Wmat}[0]{{{\bf W}}}

\newcommand{\fv}{\boldsymbol{f}}

\newcommand{\xv}{\boldsymbol{x}}
\newcommand{\yv}{\boldsymbol{y}}

\newcommand{\Thetamat}{\boldsymbol{\Theta}}

\newcommand{\Phimat}{\boldsymbol{\Phi}}

\newcommand{\phiv}{\boldsymbol{\phi}}

\newcommand{\adhoc}{{\em ad~hoc} }

\newcommand{\inv}{^{-1}}
\newcommand{\etal}{{\em et al.}}
\newcommand{\eg}{{\em e.g.}}
\newcommand{\etc}{{\em etc.}}
\newcommand{\ie}{{\em i.e.}}

\begin{document}
\setlength{\parskip}{.02in}
	
	
\title{Image Compression Based on Compressive Sensing: End-to-End Comparison with JPEG}

\author{
		\authorblockN{Xin Yuan, {\em Senior Member, IEEE} and Raziel Haimi-Cohen, {\em Senior Member, IEEE}} \\
		\thanks{Xin Yuan with Bell Labs, 600 Mountain Avenue, Murray Hill, NJ, 07974, USA, xyuan@bell-labs.com. 
			Raziel Haimi-Cohen is with Verizon Labs, 180 Washington Valley Rd, Bedminster, NJ 07921, USA, Raziel.Haimi-Cohen@verizon.com.
		The MATLAB code used to generate the results and more results are available at~\cite{CSvsJPEG_webpage}.}
}

\maketitle
\begin{abstract}
We present an end-to-end image compression system based on compressive sensing. 
The presented system integrates the conventional scheme of compressive sampling (on the entire image) and reconstruction with quantization and entropy coding.  The compression performance, in terms of decoded image quality versus data rate, is shown to be comparable with JPEG and significantly better at the low rate range. 
We study the parameters that influence the system performance, including ($i$) the choice of sensing matrix, ($ii$) the trade-off  between quantization and compression ratio, and ($iii$) the reconstruction algorithms.
We propose an effective method to select, among all possible combinations of quantization step and compression ratio, the ones that yield the near-best quality at any given bit rate.
Furthermore, our proposed image compression system can be directly used in the compressive sensing camera, \eg, the single pixel camera, to construct a {\em hardware compressive sampling} system.
\end{abstract}
	

\section{Introduction \label{Sec:intro}}
Multimedia is emerging due to advanced developments of social networks and the mobile
Internet. Billions of image and video resources are transmitted through miscellaneous ways on the Internet everyday. 
Limited by the bandwidth and storage, image and video compression still play a pivotal role in multimedia.
Meanwhile, Compressive Sensing (CS)~\cite{Candes06ITT,Donoho06ITT,Elad10_sparse} has been proposed for more than a decade as a method for dimensionality reduction of signals that are known to be sparse or compressible in a specific basis representation. 
By ``sparse”" we mean that most coefficients of the representation are zero; ``compressible" indicates that the magnitude of the coefficients decays quickly according to a power law, hence the signal can be well approximated by a sparse signal. 
In the CS paradigm, the signal is projected onto a low-dimension space, resulting in a {\em measurements vector}. For a sparse signal, it is possible to exactly reconstruct it from the measurements vector. If the measurements are noisy or the signal is not sparse but compressible, the reconstruction yields an approximation to the original signal. Natural images are inherently compressible in the frequency or wavelet domain and therefore suitable for CS. 
The past few years saw  impressive progress in this field with new reconstruction algorithms~\cite{Dong14TIP,Mertzler14Denoising} achieving  better  reconstructed image quality at a lower  compression ratio\footnote{In image processing, ``compression ratio"  often denotes the ratio between the number of bits in the original image and the number of bits in the encoded image. However, in this paper we use ``compression ratio" in the CS meaning of ratio between number of pixels and number of measurements.}, {\em i.e.},  the ratio between the dimension of the measurements vector and the number of pixels in the original image, denoted by CSr.  These algorithms go beyond sparsity and leverage other properties of natural images, such as having low rank~\cite{Dong14TIP} or being capable of denoising~\cite{Mertzler14Denoising}.
CS has been used in image and video compression for multimedia~\cite{Yuan16TMM_privacyCS,Song17TMM_CScloud,Chen18TMM_BCS,Liu16TMM_PCA,Liu14TMM_BitVideo,Zhang16TMM_BiCS,Deng12TMM}.

Encouraged by these achievements, we set out to create an end-to-end image compression system based on CS. In itself CS is {\em not} a complete signal compression system because its ``compressed signal", the measurements vector, 
is an array of real numbers rather than a sequence of bits or bytes. {Thus, in order to build a pratical compression system we added a {\em quantization stage}, in which the real-valued measurements are mapped into codewords from a finite codebook, and a {\em lossless coding stage}, which converts the  codeword sequence into a byte sequence, using entropy coding.} 

\subsection{Related Work}
Goyal \etal\cite{Goyal08SPM} applied an information-theoretic approach to assess the effectiveness of a CS-based compression system for sparse signals  $\xv \in {\mathbb R}^{N}$, with only $K$ non-zero entries, a.k.a., $K$-sparse. Their benchmark was the ``baseline" method, where the coded bit sequence consisted of the sparsity pattern (\ie, a specification of the indices of the non-zero entries in $\xv$) and the quantized and coded non-zero entries. They showed that the rate-distortion functions of a CS-based compression system are considerably worse than those of the baseline method for two reasons. First, the number of measurements $M$ required by CS to recover $\xv$ is several times larger than $K$, $M/K \ge  \log(N/K)$, and the number of bits needed to represent the additional $M-K$ quantized variables exceeds the number of bits needed to specify the sparsity pattern, especially when the number of bits per measurement is high. Second, the quantization noise in the baseline method is proportional to $K$, whereas in CS with a random sensing matrix, it is proportional to $M$. Goyal \etal ~suggested that the use of distributed lossless coding and entropy-coded dithered quantization might potentially alleviate those problems, but the complexity added by those methods would probably make them impractical. 
Despite this pessimistic outlook, the research of the effect of quantization on CS measurements received significant attention in the last few years. The trade-off between the number of measurements and quantization accuracy was investigated in~\cite{Laska12_TSP_bitchange,LASKA11_ACHA}, and methods of accounting for the quantization effect (including the extreme case of a 1-bit quantizer) in the reconstruction algorithm were studied in~\cite{Zymnis10SPL,Laska12_TSP_bitchange,Laska11_TSP_1bit,Dai09_ISIT,Dai09_ITW,Baig10_ICT,Venkatraman09_ICASSP}.  
However this significant body of research was of limited value for our purposes. First, these works assumed a random, or a structurally random sensing matrix, while the sensing matrices suitable for our purposes could be different (see Sec.~\ref{Sec:SM_Mea}). Second, most of these works did not assume any {lossless} coder and therefore did not consider the resulting bit rates. Those that {did consider a lossless} coder did not study the interaction between the quantizer and the {lossless} coder and the trade-offs in their design.

In the signal compression field, a ``compression system" converts the signal to a bit stream at the encoder and reconstructs the signal from the bit stream at the decoder. The performance of such a system is measured in terms of the {\em reconstruction error vs. bit rate}. There are various papers reporting the effect of measurement quantization on the reconstruction error~\cite{Laska11_TSP_1bit,Deng12TMM,Dong14TIP}; however, these studies present their results in terms of reconstruction error {\em{vs.}} compression ratio, {rather than {\em{vs.}} bit rate}. This is because they do not consider the {quantization and lossless} coding aspects, that is, converting the sequence of quantization codewords into a low rate bit stream. This is the main reason there has been no comparison of performance between CS and classical image compression methods such as JPEG and JPEG2000 --- without a bit rate there is no way of meaningful comparison.
While we were preparing this work for publication we came across a deep learning inspired paper by Cui {\em et al.}~\cite{Cui18MM} (among other deep learning based methods), which also describes an end-to-end image compression system based on CS\footnote{The description in~\cite{Cui18MM} was not detailed enough to allow replication and comparison in our lab, but we highlight the main similarities and differences in the system design relative to our work.
1) Cui {\em et al.}, like us, recognize that a random, or structurally random sensing matrix will not perform well in this application, and they use a neural network to train a sensing matrix. The histogram of measurements shows that most of the measurement energy is concentrated in a just a few measurements, similar to the effect of classical transforms like DCT or WHT, but without the computational advantages of the fast transforms.
2) Cui {\em et al.}'s system, like ours, uses uniform quantization and arithmetic coding. Their system does seem to jointly optimize the compression ratio and the quantization step, but it has a neural network sub-module for improving the dequantization. The use of this module results in a modest improvement in PSNR about 0.4 dB.
3) The reconstruction in Cui {\em et al.}'s system is done by a neural network, which is trained jointly with the sensing matrix and the dequantization correction sub-network. Consequently, unlike our system, their encoder and decoder seem to be tightly coupled.
4) Cui {\em et al.}'s system works on blocks of 32x32 samples, unlike our system which operates on the whole image. A second decoding stage (\ie a second neural network) is used to eliminate blocking artifacts. This blocking may be because full frame training would require prohibitively large amount of training materials and computational resources.
5) Cui {\em et al.} compare their results with other CS-based compression systems to which they added fixed length {lossless} coding. There is no comparison with standard image compression methods like JPEG or JPEG2000 as performed in this paper.}.


\subsection{JPEG, JPEG2000 and Uniqueness of CS-based Image Compression}
Since its introduction in 1992, JPEG~\cite{JPEG1994} has been one of the most popular image compression methods in use. JPEG2000~\cite{JPEG2000}, which was released ten years later, has  superior compression properties at a wider bit rate/quality range. A brief overview of these standards is given in  Sec.~\ref{Sec:CompareJPEG}, where their architectures are compared with the proposed framework, dubbed compressive sensing based image compression system (CSbIC). 

{HEIF~\cite{Hannuksela2015TheHE,Lainema2016HEVCSI} (the still-image version of HEVC) and WebP~\cite{Ginesu2012ObjectiveAO} are recently introduced image compression standards based on intra-frame prediction, which claim to have better compression than JPEG.}
Another recent approach is sparse representation image coding, after JPEG2000, where the wavelet basis is replaced by a learned, over-complete dictionary with constraints on the variance of the representation coefficients~\cite{Xu14TCSVT,Zhang17DCC,Zhang18TIP}. However, direct comparison of dictionary learning based methods with other methods is difficult, since the dictionaries are learned for a specific class of images.

At the time of its introduction, the higher complexity of JPEG2000 was an impediment for its adoption, but as computing capabilities improved, this was no longer an issue, and this is true for the aforementioned compression enhancements. Nevertheless, JPEG remains the image compression tool of choice in a wide range of applications. 
{It appears that for most conventional applications (web, professional and amateur photography, publishing, \etc), the advantages offered by the more advanced compression schemes, were not significant enough to justify migration to a new standard. This raises the question whether there is  need for yet another image compression scheme}. In response, a compression scheme based on CS has some unique properties which are radically different from those of any conventional signal compression scheme. 

\begin{itemize}
	\item 
	The encoder is of much lower complexity than the decoder. In fact, a significant part of the encoder processing, namely the measurements generation, can be done in the analog domain via a single-pixel camera or lensless camera~\cite{Duarte08SPM,Huang13ICIP,Yuan18OE}. 
	\item 
	The decoder is not fully defined by the encoder. Different reconstruction algorithms can be applied to the same measurements vector, so as more prior information becomes available the reconstruction algorithm can be improved to take it into account~\cite{Dong14TIP,Mertzler14Denoising}.
	\item 
 {The loss of a small fraction of the measurements generally results in only a minor degradation in reconstruction quality~\cite{Laska11_TSP_1bit,LASKA11_ACHA}}. This may be used to achieve robustness to channel impairments without the usual overhead of error protection.	
\end{itemize}

These unique properties suggest that CSbIC would be very useful in {new, non-conventional applications such as media sensor networks, where the sensors need to be inexpensive and have low energy consumption and the network operates at low power and is usually unreliable. In contrast, the control center has ample processing power for decoding, and  much prior information may be gathered while the system is running.} Therefore, CSbIC is suitable for a class of applications that are different from those that JPEG and JPEG2000 were designed for.  

\begin{figure}[!htbp]
	\begin{center}
		\includegraphics[width=1\linewidth]{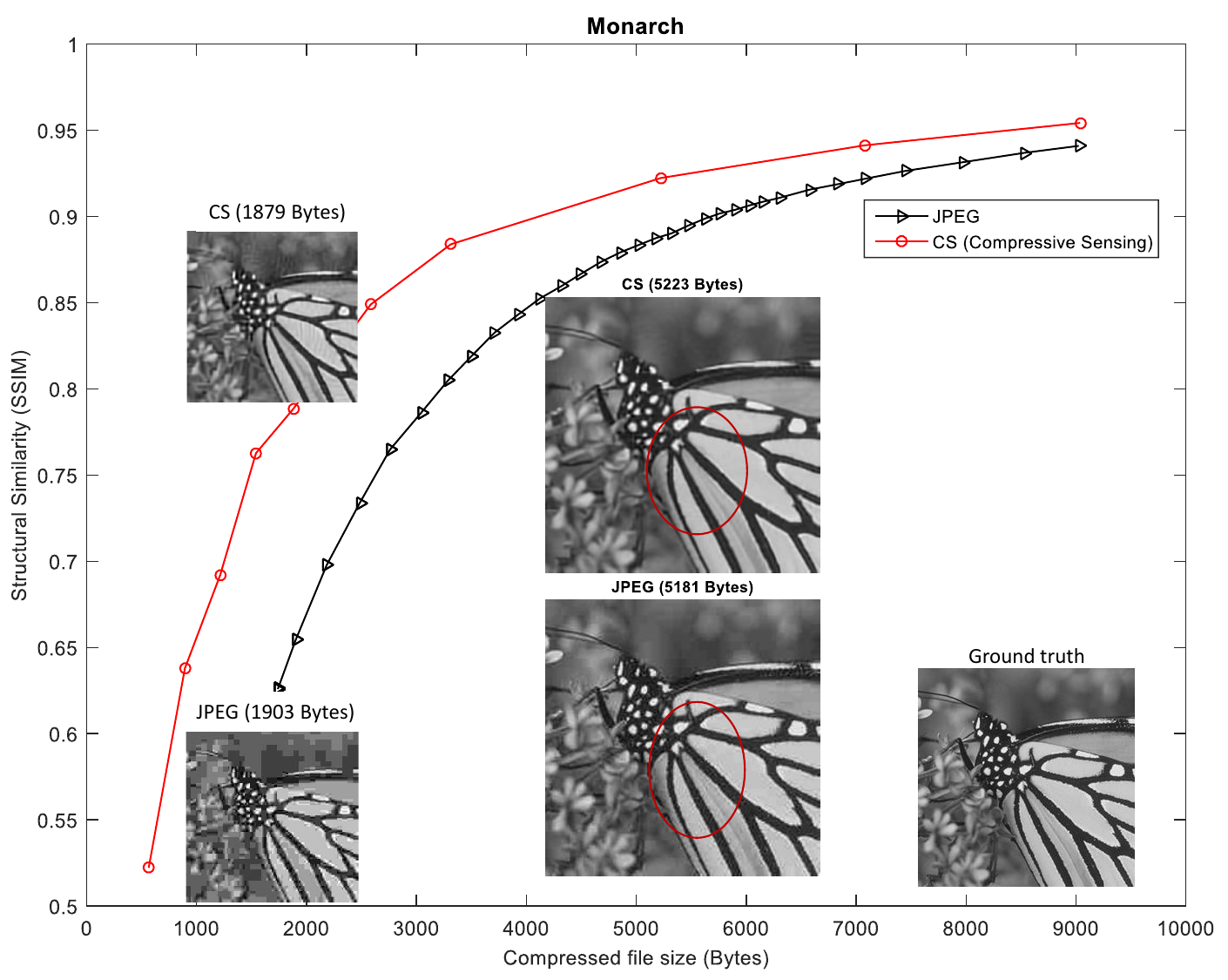}
	\end{center}
	\vspace{-3mm}
	\caption{Comparison of JPEG and CS-based compression of the ``Monarch'' image. Curves of structural similarity (SSIM)~\cite{Wang04imagequality} vs. compressed file size, as well as decoded images of similar file size are shown. Please zoom-in on red circles for details. }
	\label{fig:monarch_front}
\end{figure}

\subsection{Contributions and Organization of This Work}
This paper makes the following contributions:
\begin{itemize}
	\item [$i$)] 
	We provide an end-to-end architecture for a CSbIC system, including compressive sensing, {quantization, lossless coding} and a mechanism to adjust the system parameters to control data rate and reconstruction quality.
	\item [$ii$)]
	We address the theoretical issues raised by Goyal \etal\cite{Goyal08SPM} in a practical way by using {\em domain-specific} knowledge: (i) we employ reconstruction algorithms that do not rely solely on sparsity but also on other properties of natural images, and (ii) we use deterministic sensing matrices that are known to be effective for such signals.
	\item [$iii)$]
	Having an end-to-end system enables us to {benchmark its performance in terms of quality versus data rate. In order to give the reader a sense of CSbIC compression performance, we compare it with the popular JPEG and JPEG2000 standards. Our goal is not to propose CSbIC as a competitor to these and other standards such as HEIF or WebP. As we pointed out, in conventional applications for which these standards were designed for, the image compression needs are already well met. Rather, the purpose of the comparison to JPEG and JPEG2000 is to give the reader comparative levels of compression that can be expected.} We show that our CSbIC system is on-par with JPEG, with a clear advantage in the low data rate range. Please refer to Fig.~\ref{fig:monarch_front} as an example.
	\item [$iv$)]
	We describe various design choices in each of the system components and study the effects that these choices have on the overall system performance.
\end{itemize}
The system that we describe is far from being fully optimized, and throughout the paper we point out where further improvements may be made. Nevertheless, even at this preliminary stage the performance of our system makes it a viable alternative to the conventional methods (refer to Figs~\ref{fig:SSIM_8img_3algo}-\ref{fig:imges_cs_JPEG} for comparison with JPEG).

The rest of this paper is organized as follows.
Sec.~\ref{Sec:Arch} describes the system architecture and discusses the design choices in each component. 
Sec.~\ref{Sec:rec} provides the general framework of reconstruction algorithms. 
Sec.~\ref{Sec:perf_mea} presents the results of our performance testing and Sec.~\ref{Sec:dis}  discusses the implications of our results.

%

\begin{figure*}[tbp!]
	\centering
	\includegraphics[width=1\textwidth]{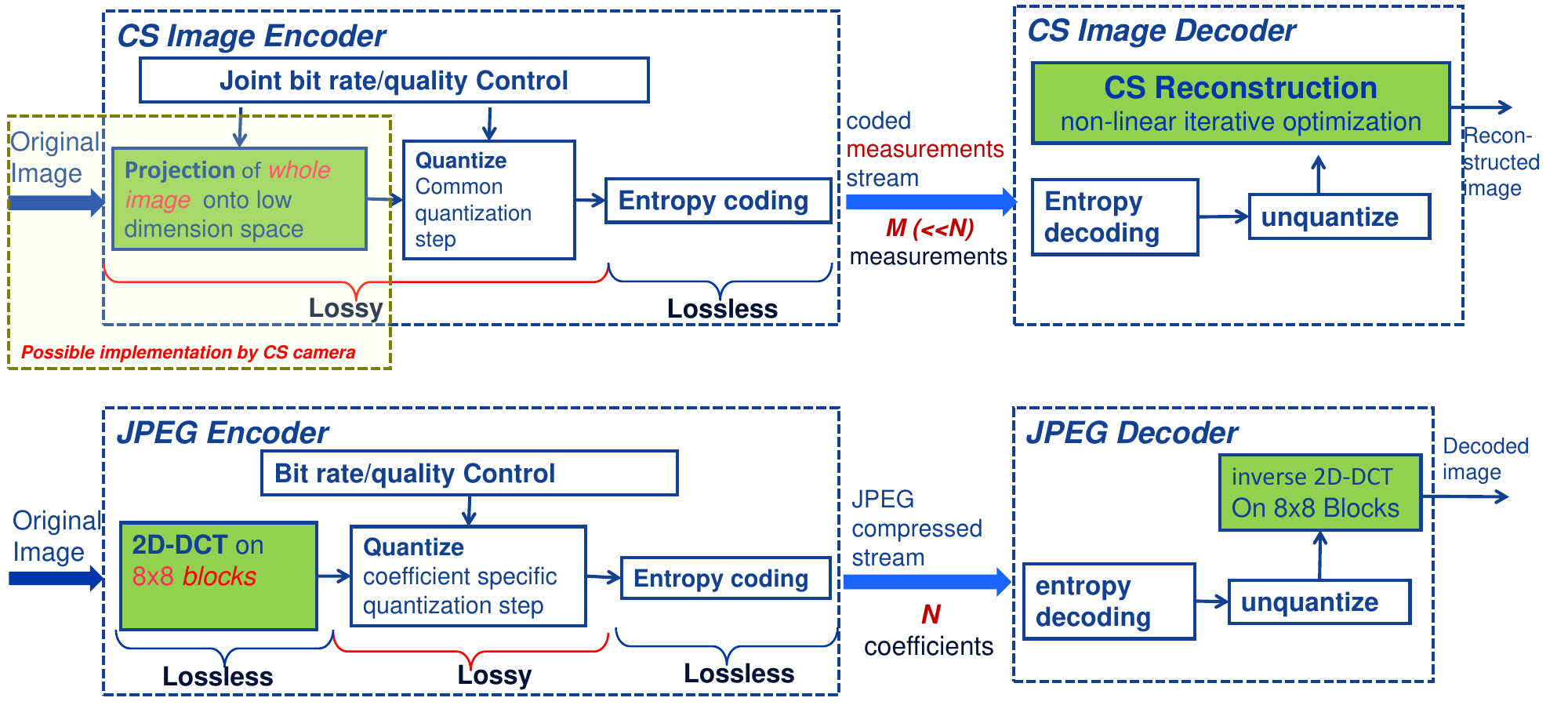}
	\vspace{-3mm}
	\caption{Image compression architecture comparison between proposed CSbIC (top) and JPEG (bottom). }
	\label{Fig:Arch}
	\vspace{-3mm}
\end{figure*}
\section{System Architecture \label{Sec:Arch}}
A diagram of the system architecture is given in Fig.~\ref{Fig:Arch}. Each of the encoding steps is matched by a corresponding decoding step (in reverse order), with the exception of the bit rate/quality control block, which appears only in the encoder. In the following we present a detailed description of each of those processing steps.
The upper part of Fig.~\ref{Fig:Arch} shows the corresponding block diagram of the JPEG encoder and decoder. 
\begin{itemize}
	\item In JPEG the first stage of the encoder (2D-DCT transform on $8\times8$ blocks) is {\em lossless} -- the output is of the same dimension, $N$, as the input. The same is true for JPEG2000, although the transform is different (it is a wavelet transform on the whole image or tile). Therefore the quantizer and lossless coder have to process  an $N$-dimensional signal, and their computational workload is about the same regardless of the compression level.
	\item  In CSbIC, the first stage (projection of whole image onto low dimension space) is {\em lossy} -- the output of the first stage is of dimension $M\ll N$, and hence the computational workload of the quantizer and lossless coder is much lower, proportional to $M$. Furthermore, if we do not use a fast transform in the first stage, it becomes a simple multiplication of a vector with dimension $N$ by a matrix of dimension $M\times N$, with $O(MN)$ complexity. This multiplication can be performed in the analog domain, using a CS camera, or it can be implemented by efficient dedicated hardware. In either case, the complexity of the first stage is also proportional to $M$. As we will see in Sec.~\ref{Sec:QualityCon}, when the joint bit-rate/quality control block decreases the quality, it does so in part by decreasing $M$. Therefore, in addition to the usual trade-off between quality and bit rate, CSbIC offers a trade-off between quality and encoder computational complexity.
\end{itemize} 
As we will see later (Sec.~\ref{Sec:results_algo}), the choice of reconstruction algorithm offers a similar trade-off in the decoder. Thus, CSbIC allows dynamic adjustment of quality based on available computational resources.

\subsection{Sensing Matrix and Measurements Generation \label{Sec:SM_Mea}}
We consider monochromatic images of $N_v\times N_h$ pixels. The pixels of the input image are organized column by column as a pixel vector  $\xv \in {\mathbb R}^{N}$ ($N = N_vN_h $). The pixel vector $\xv$ is multiplied by a sensing matrix $\Phimat\in {\mathbb R}^{M\times N}$, $M\ll N$, yielding the measurements vector
\begin{equation}
{ \yv = \Phimat \xv}. \label{Eq:yPhix}
\end{equation}
$\Phimat$ is a large matrix, even for small images. Therefore, for practical reasons, $\Phimat$ is never stored, and the operation \eqref{Eq:yPhix} is implemented by a fast transform. It is well known that a 2-dimensional discrete cosine transform (2D-DCT) is effective in decorrelating an image, and most of the energy of the image is concentrated in the low frequency coefficients. Recently, new  sensing matrices were introduced that leverage this property~\cite{Romberg08SPM,ahn2016compressive}. These matrices are {\em not} incoherent with the common sparsity bases of natural images, as classical CS theory would require for guaranteeing robust reconstruction~\cite{Candes06ITT,Elad10_sparse}, but generally they perform better than the classical sensing matrices in our application of  image compression. For example, \eqref{Eq:yPhix} can be implemented by performing a 2D-DCT on the image pixels and then reordering the resulting coefficients in a ``zig-zag" order (similar to the one used in JPEG encoding), and then selecting the first $M$ low-frequency coefficients~\cite{Romberg08SPM}.

In some applications, such as the single-pixel camera and lensless camera~\cite{Duarte08SPM,Huang13ICIP,Yuan15Lensless,Yuan16SJ,Yuan18OE}, a binary-valued matrix, \ie, a matrix whose entries are only $\pm1$ (or $\{0,1\}$), is more suitable for hardware implementation. In this case we approximate the 2D frequency decomposition by using a 2D Walsh-Hadamard transform (2D-WHT)~\cite{Pratt69_IEEE}. Let 
\begin{equation}
{  \Wmat = \Wmat_h \otimes \Wmat_v}, \label{Eq:2DDWT}
\end{equation} 
where $\otimes$ denotes the Kronecker product~\cite{Duarte12_KroCS} and $\Wmat_h, \Wmat_v$ are Walsh-Hadamard matrices in sequence order~\cite{Fino76_TC_WHT}. 
Similar to the 2D-DCT case, the selected measurements are the first $M$ coefficients of the zig-zag ordered entries of $\Wmat\xv$, which can also be computed numerically in an efficient way.

We can get the CS theoretical guarantee for successful reconstruction, w.h.p. (with high probability), by replacing the deterministic matrices described above with random matrices whose entries are  independent, identically distributed (IID) random variables (RVs) with Gaussian or Rademacher distributions~\cite{cs_Candes06randomProj,Candes05_LinearP}. These matrices are universal, \ie, w.h.p. they are incoherent with any given sparsity basis. Furthermore, the measurements generated by those random matrices are mutually independent and asymptotically normally distributed, which is helpful in the quantization and coding design. Such fully random matrices do not allow fast transform implementation of \eqref{Eq:yPhix}. But similar desired properties and performance guarantees were shown for structurally random matrices (SRM)~\cite{Haimi-CohenL16_SP,Do12_SRM,Do08_ICASSP} where $\Phimat \xv$ is obtained  by applying a randomly selected permutation to $\xv$, computing a fast transform, on the permuted vector, and randomly selecting $M$ of the transform coefficients as measurements (the DC coefficient is always selected). We denote these matrices SRM-DCT and SRM-WHT, respectively.

\subsection{Quantization \label{Sec:Quant}}
The quantizer maps the measurements vector $\yv$ into a finite sequence of codewords taken from a finite codebook $\cal C$ and the dequantizer maps the codewords sequence into a measurements vector that is an approximation of the original measurements vector. If the sensing matrix is deterministic, the measurements are highly uncorrelated; if it is a SRM the measurements are nearly independent. Hence the advantage of vector quantization~\cite{Gersho1991} over scalar quantization is small and does not justify its added complexity~\cite{Dai09_ISIT}. Therefore, we consider a scalar quantizer that maps each measurement 
$\{y_i\}_{i=1}^M$ to a codeword $q_i= Q_i(y_i) \in {\cal C}$, where $Q_i: {\cal R}\rightarrow {\cal C}$ is the quantizer of the $i^{th}$ measurement. In this work we use the same quantizer for all measurements, and hence in the following we omit the subscript $i$ from $Q_i$.  

The simplest scalar quantizer is the uniform quantizer. We select the ``mid-tread" type, defined by
\begin{align}
Q(y) &{\textstyle \stackrel{\rm def}{=}  \max(-L, \min(L, \tilde{Q}(y)))}, \\
\tilde{Q}(y)&{\textstyle \stackrel{\rm def}{=} \lfloor (y-\mu)/s +0.5 \rfloor}, \label{Eq:til_Q}
\end{align}
where $\lfloor y \rfloor$ denotes the largest integer not exceeding $y$, $\mu = \frac{1}{M}\sum_{i=1}^M y_i$ is the mean of the measurements, $s$ is the quantizer's step, $\tilde{Q}(y)$ is the unclipped  quantized value, and $L$ is a positive integer that determines the range $s(L-0.5)$ of the actual quantizer $Q(y)$. Consequently there are $2L+1$ codewords, ${\cal C}=\{-L, \cdots, L\}$.
Since the distribution of the measurements is highly non-uniform, the codewords distribution is also not uniform; hence in order to represent codewords effectively by a bit sequence we need to use variable length coding (VLC) in the {lossless} coder. 
A uniform quantizer with fixed length coding (FLC) with little data rate penalty is used due to its simplicity.
Another reason to use a uniform quantizer is that the reconstruction may be sensitive to the presence of even a few measurements with large errors~\cite{LASKA11_ACHA}, which is often the case with non-uniform quantizers.

If $Q(y) = {c}$ and $|{c}|<L$, we define the dequantizer by
\begin{equation}
{\textstyle Q\inv({c}) = { c} s + \mu}, \label{Eq:Dequant}
\end{equation}
hence the quantization error is bounded by
\begin{equation}
{\textstyle |y - Q\inv(Q(y)) |\le 0.5s}.
\end{equation}
If $|c| =L$, the quantized measurement is {\em saturated} and the quantization error cannot be bounded.  
Even a small number of saturated measurements can cause severe quality degradation unless they are specially handled in the reconstruction~\cite{LASKA11_ACHA}. The simplest way to do it is by not using the saturated measurements at all; attempts to modify the reconstruction algorithm to use those measurements showed little gain over simply discarding them. Another option (not considered in~\cite{LASKA11_ACHA}) is to code the value of $\tilde{Q}(y)$ for each saturated measurement $y$ in some \adhoc manner and transmit it as additional information. In both cases saturated measurements incur a penalty, either in the form of transmitted codewords that are not used, or as \adhoc transmission of $\tilde{Q}(y)$. Therefore, 
we select $L$ large enough to make saturation~\cite{LASKA11_ACHA} a rare event. In fact, $L$~can be set sufficiently large to eliminate saturation completely, but a very large codebook may have an adverse effect on {lossless} coding (see Sec.~\ref{Sec:CodeHist}). We found that a good trade-off is to select  $L$ so that quantizer's range $s(L-0.5)$ is about 4 standard deviations of the measurements. However, the system is not very sensitive to this parameter --- performance does not change much if the range is 3 or 6 standard deviations. 

With all the sensing matrices considered, the first measurement $y_1$ is the DC coefficient, which is the sum of all pixels in the image; $y_1$ is much larger than the other measurements and is always saturated, therefore it requires special handling: $y_1$ is excluded when calculating the mean ($\mu$), and the standard deviation of the measurements, and $Q(y_1)=L$ is not included in the quantized measurement. Instead, $\tilde{Q}(y_1)$ is coded in an \adhoc fashion and transmitted separately.
Unless the quantization is very coarse, its effect on the measurements can be modeled as adding white noise, uniformly distributed in $[-s/2,s/2]$, which is uncorrelated with the measurements~\cite{Gersho1991}. Hence the variance of the quantization noise in each measurement is
\begin{equation}
{\textstyle \sigma_Q^2 = s^2/12}.
\end{equation}
The integral pixel values are generally obtained by sampling an analog signal and rounding the samples values to the nearest integer. Hence the pixel values contain digitization noise with variance of $1/12$. This noise appears in the measurement $y_j$ with variance 
\begin{equation}
{\textstyle \sigma_D^2 = \|\phiv_j\|_2^2/12}, \label{Eq:DigitizationSigma}
\end{equation}
where $\phiv_j$ is the $j^{th}$ row of $\Phimat$. In the sensing matrices that we consider $\|\phiv_j\|_2$ is constant, $\|\phiv_j\|_2=\|\Phimat\|_2$. Clearly, there is no point in spending bits to accurately represent the digitization noise, hence we need to have $\sigma_Q \geq \sigma_D$ and consequently $s \geq \|\Phimat\|_2$. 

\subsection{Image Quality Control \label{Sec:QualityCon}}
As mentioned earlier, the performance of a compression system is measured in terms of reconstruction quality {\em vs.} bit rate. 
Usually, both of these quantities are determined by a single parameter.  By varying this parameter we obtain the receiver operator curve (ROC), which depicts reconstruction quality against bit rate. 
However, in our case, both the coded image size and the quality of the reconstructed image depend on two parameters, the compression ratio $R$ and the quantizer step size $s$. One can get the same coded image size with various combinations of these two parameters, but the reconstruction quality varies significantly among those parameter combinations.  If the encoded image size is constrained not to exceed $b$, then the highest possible reconstruction quality $Z^*(b)$ is given by
\begin{equation}
{Z^*(b) = \arg\max_{R,s} Z(R,s),  ~~ {\text {subject to}}~~ B(R,s) \le b,}  \label{Eq:Zb}
\end{equation}
where $B(R,s)$ and $Z(R,s)$ are the coded image size and the reconstructed image quality (according to some specified quality criterion) respectively, as functions of the compression ratio and the step size. 
For each value of $b$, there is an optimal pair of compression ratio and quantization step, $(R^*(b), s^*(b))$, which is the solution of  the optimization problem \eqref{Eq:Zb}.  
However, getting an analytical solution to this equation, even approximately, turns out to be an extremely difficult problem\footnote{The joint compressive sensing (using sensing matrix to obtain measurements) and data compression (to quantize the measurements) is an unsolved problem in theory to our best knowledge.
The most related work can be found in~\cite{Kipnis17ISIT}; unfortunately, no closed-form solution has been shown.
Most recently, Goldsmith {\em el al.}~\cite{Kipnis16TIT,Kipnis16ADC} performed some research on the distortion limits of analog-to-digital compression, which can inspire some analysis for this joint analysis of compressive sensing and data compression. However, significant gaps still exist to derive any explicit formula.}. 
It is relatively simple to obtain an approximate expression for $B(R,s)$\footnote{Since we are using entropy coding, we may approximate the bit rate by the entropy of the measurements, 
\begin{equation}
{\textstyle B(R,s) \approx \sum_{j=1}^M H_s(y_i) \approx \sum_{j=1}^M H(y_j) -M \log_2 s},
\end{equation}	
where $H_s(y_j)$ is the entropy of the measurement $y_j$ quantized with a step size $s$, $H(y_j)$ is the differential entropy of $y_j$, and the second inequality is based on the well-known approximation (for small step size), $H(y_j) = H_S(y_j) -\log_2 s$. $H(y_j)$ can be computed by estimating the distribution $y_j$ over a large number of images, or in the case of stochastic sensing matrices, since the measurements are known to be approximately Gaussian and identically distributed, one could get an analytic expression for $H(y_j)$.}, but unfortunately, there is no closed form expression for $Z(R,s)$. 
The CS literature provides several upper bounds for the mean square reconstruction error~\cite{Candes11RIPless}, but using any of these upper bounds as a substitute to the actual mean square error is a gross oversimplification and the formulas derived for $(R^*(b), s^*(b))$ from this ``approximation" have no resemblance to the empirical results.  Furthermore, the mean square error does not reflect image quality as perceived by the human visual system (this issue is elaborated in Sec.~\ref{Sec:result_jointcontrol}); hence even if we had an expression for $Z(R,s)$ in the mean square error sense, it would not help us in solving~\eqref{Eq:Zb} when the objective is to maximize the perceived quality.

Since an analytic solution is not available, we set out to search for an empirical one. 
For eight widely used images (shown in Fig.~\ref{fig:SSIM_8img_3algo}), we plotted curves of $Z(R,s)$ against $B(R,s)$, where $R$ was kept fixed and $s$ was the independent variable. 
{Plotting these curves in the same graph, for different values of $R$ (see the example in Fig.~\ref{Fig:Joint_control}), made it clear that each value of $R$ is optimal only for a particular quantization step, $s^*(R)$, that is $(R,s^*(R))$ are the solution to~\eqref{Eq:Zb} for a particular value of $b$.
For $s>s^*(R)$, we could find $R'<R$ such that $B(R',s) = B(R,s^*(R))$ but $Z(R',s)>Z(R,s^*(R))$, and similarly, for $s<s^*(R)$, we could find $R''>R$ such that $B(R'',s) = B(R,s^*(R))$ but $Z(R'',s)>Z(R,s^*(R))$. Therefore, for each image the function $s^*(R)$ was monotonically decreasing, but it was different for different images. A closer examination of the data revealed that in each image $R$ and $s^*(R)$ were approximately inversely proportional to each other, with a proportion constant of the same order of magnitude as $\|\Phimat\|_2$. This could be expressed by:}
\begin{equation}
 R~s^*(R) = C \|\Phimat\|_2, \label{Eq:RSN}
\end{equation}
where $C$ is a constant whose value depends on the type of sensing matrix and varies from image to image, {but generally is between 1 to 10.}
Furthermore, we found that using $C = 2.0$ is a good choice, albeit suboptimal, for all images.  To confirm these findings we ran our tests on 200 images from the Berkeley Segmentation Dataset and Benchmark dataset, \ie, BSDS300~\cite{MartinFTM01} and verified that setting $(R,s)$ using~\eqref{Eq:RSN} with $C = 2.0$ is a generally good choice for all natural images. Thus in our tests $s^*(R)$ is determined using~\eqref{Eq:RSN}, with $C=2.0$. The resulting quantization step is sufficiently fine to enable modeling the quantization noise as uncorrelated, uniformly distributed white noise.  

It should be emphasized that our choice of $C=2.0$ is not optimal for many images, and Eq.~\eqref{Eq:RSN} itself is an empirical approximation of a very complex relationship between the optimal compression ratio and optimal step size. Consequently, one could often get better results than those presented in this paper by manually adjusting $s^*(R)$ for a given image. Therefore, finding a better practical solution for the optimization problem~\eqref{Eq:Zb} could lead to significant performance improvement in our system.

\subsection{Lossless Coding \label{Sec:Lossless}}
The lossless encoder encodes the codeword sequence generated by the quantizer as well as some miscellaneous information (\eg, $\mu, s$, and the \adhoc representation of saturated measurements) as a bit sequence. The lossless decoder exactly decodes the codeword sequence and the miscellaneous information from the bit sequence.  
\subsubsection{Coded Numbers Format \label{Sec:Num_Format}}
Various types of numbers are coded by the lossless encoder (and decoded by the lossless decoder). Each type is encoded in a different way:

{\bf {Unbounded signed or unsigned integers}} are integers whose maximal magnitude is not known in advance.  They are represented by byte sequences, where the most significant bit (MSB) in each byte is a continuation bit. It is clear in the last byte of the sequence and set in all other bytes. The rest of the bits are payload bits which represent the integer. The number of bytes is the minimal number that has enough payload bits to fully represent the unsigned or unsigned integer.

{\bf {Real numbers}}, which are natively stored in single or double precision floating point format~\cite{Langdon79arithmeticcoding} are coded as pairs of unbounded signed integers representing the mantissas and the exponents in the floating point format.

{\bf {Bit arrays}} are zero padded to a length which is a multiple of 8 and coded as a sequence of bytes, 8 bits per bytes.

{\bf {Bounded unsigned integer arrays}} are arrays of unsigned integers, each of which may be represented by a fixed number of bits. An array of $n$ integers, each of which can be represented by $b$ bits, is encoded as a bit array of $bn$ bits.

Each of these number formats can be easily decoded. Note that these formats are byte aligned for simpler implementation.

\subsubsection{Entropy Coding \label{Sec:EntrpCdng}}
The codewords $\pm L$ represent saturated measurements. 
We merge these two labels into a single label $L$, thus we have $2L$ codewords:  $  -L+1,\dots,L $. 

Let $p_c, c\in {\cal C}$ be the probability of a measurement to be quantized to $c$. If the codewords $\{Q(y_i)\}_{i=1}^M$ are IID random variables, then a tight lower bound on the expected number of bits required to transmit these codewords is the entropy rate:
\begin{equation}
{\textstyle H\stackrel{\rm def}{=}-M\sum_{c\in {\cal C}} p_c \log_2 p_c.} \label{Eq:H_codeword}
\end{equation}
Arithmetic coding (AC)~\cite{Langdon79arithmeticcoding} represents the codeword sequence by a bit sequence, the length of which can get arbitrarily close to the entropy rate for a large $M$. We use AC to encode the codewords sequence. 

Since the probabilities $p_c, c\in{\cal C}$ are not known {\em a priori}, they need to be estimated and sent to the receiver, in addition to the AC bit sequence. These probability estimates can be obtained in two ways: For SRMs, the measurements are approximately normally distributed~\cite{Haimi-CohenL16_SP}, hence for $|c|<L$, $p_c$ is the normal probability of the quantization intervals $[cs+\mu-0.5s, cs+\mu+0.5s)$, and $p_L = 1-\sum_{|c|<L} p_c$. Thus, all that needs to be sent to the receiver is the estimated standard deviation of the measurements, which is coded as a real number. For deterministic sensing matrices, it is necessary to compute a histogram of the quantized measurements sequence, use it to determine the probabilities, and then code the histogram and send it to the receiver along with the AC bit sequence. Sending the histogram is an overhead, but it is small in comparison to the gain achieved by arithmetic coding. 

In natural images the magnitudes of the coefficients of 2D-DCT or 2D-WHT decay quickly, hence measurements generated using a deterministic sensing matrix are not identically distributed, which violates the assumptions under which AC is asymptotically optimal. In order to handle this problem  we partition the codeword sequence into sections, and for each section we compute a histogram and an AC sequence separately. This, of course, makes the overhead of coding the histograms significant. In the following we describe how the histograms are coded efficiently and how to select a locally optimal partition of the codewords sequence.

\subsubsection{Coding of Histograms \label{Sec:CodeHist}}
In order to be efficient, the code of histograms of short codeword sequences should be short as well. Fortunately, such histograms often have many zero counts, which can be used for efficient coding. A histogram is coded in one of three ways:

{\bf {Full histogram:}} A sequence of $2L$ unbounded unsigned integers, containing the counts for each codeword.  This method is effective when most counts are non-zero.

{\bf {Flagged histogram:}} A bit array of $2L$ bits indicates for each codeword whether the corresponding count is non-zero, and a sequence of unbounded unsigned integers contains the non-zero counts. This method is effective when a significant share of the counts is zero.

{\bf {Indexed histogram:}} A bounded integer indicates the number of non-zero counts, an array of bounded integers contains the indices of the non-zero counts, and a sequence of unbounded unsigned integers contains the non-zero counts. This method is effective when most of the counts are zero. In the extreme case of a single non-zero count the AC bit sequence is of zero length, hence this histogram coding is effectively a run length encoding (RLE).

The histogram is coded in these three ways and the shortest code is transmitted. A 2-bit histogram format selector (HFS) indicates which representation was chosen. The HFSs of all sections of the codeword sequence are coded as a bit array. Thus, each section of the codeword sequence is represented by the HFS, the selected histogram representation and the AC bit sequence. 
{\small{
	\begin{algorithm*}[btbp!]
		\caption{Partitioning the codeword sequence into entropy coded sections.}
		\begin{algorithmic}
			\STATE 1. Initialization: 
			\begin{itemize}
				\setlength\itemsep{0em}
				\item[a.] Partition the codewords sequence $\{Q(y_j)\}_{j=1}^M$ into $J_0 = M$ sections of length 1, $\{S_j^0\}_{j=1}^{J_0}$.
				\item[b.] For each section, compute the histogram counts $h_j^0(c) \stackrel{\rm def}{=} h_{S_j^0}(c)$, $c \in {\cal C}$,
				$j=1, \dots, J_0$.
				\item[c.] Compute the coded sections' lengths using~\eqref{Eq:Ls}: 
				$l_j^0 \stackrel{\rm def}{=}{\cal L}(S_j^0), ~\forall j=1,\dots, J_0$.\\
				In this case, since all sections contain exactly one codeword, $\hat {\cal L}(S_j^0) = 0$ and ${\cal L}_H(S_j^0)$ is the same for all partitions.
				\item[d.] Let ${\cal P}_0 = \{(j,k)| 1\le j<k \le {\rm min}(J_0, j+m-1)\}$. For each pair $(j,k)\in {\cal P}_0$:
				\begin{itemize}
					\item [i.] Let $S_{j,k}^0$ be the section obtained by merging sections $S_j^0, \dots, S_k^0$ and compute the histogram resulting from this merge:
					$h_{j,k}^0(c) = \sum_{i=j}^k h_i^0(c), c\in {\cal C}$
					and compute the values of $\hat{\cal L}(S_{j,k}^0)$ using the histogram and Eq.~\eqref{Eq:Ls}.
					\item [ii.] Using the above histogram, compute the gain in bit rate obtained by merging sections $S_j^0, \dots, S_k^0$:\\
					$g_{j,k}^0 = \sum_{i=j}^k l_j^0 - {\hat L}(S_{j,k}^0)$.		
				\end{itemize}
			\end{itemize}
			\STATE 2. Let $m$ be a small positive integer. 
			\FOR{$r=0,1,\dots$ }
			\STATE a. Let $(j^*,k^*) = \arg\max_{(j,k)} g^r_{j,k}$.
			\STATE b. If $g_{j^*,k^*}^r \le 0$ break out of the loop.
			\STATE c. Merge sections $S^r_{j^*}, \dots, S^r_{k^*}$ to get partition $(r+1)$, in which
			{\footnotesize{
					\begin{align}
					J_{r+1} &= J_r - (k^* - j^*),\\
					h^{r+1}_j (c) &= \left\{\begin{array}{lc}
					h_j^r,  & 1\le j <j^*,\\
					h_{j^*,k^*}^r,  & j = j^*,\\
					h^r_{j+(k^*-j^*)}  & j^*<j \le J_r.
					\end{array}\right.~
					l_j^{r+1} = \left\{\begin{array}{lc}
					l_j^r,  & 1\le j <j^*,\\
					{\cal L}(S_{j^*,k^*}^r), & j = j^*,\\
					l^r_{j+(k^*-j^*)} & j^*<j \le J_r.
					\end{array}\right.~
					S^{r+1}_j (c) = \left\{\begin{array}{lc}
					S_j^r, &  1\le j <j^*,\\
					S_{j^*,k^*}^r,  & j = j^*,\\
					S^r_{j+(k^*-j^*)}  & j^*<j \le J_r.
					\end{array}\right.
					\end{align}}	}
			Let ${\cal P}_{r+1} =\{(j,k)|1\le j<k\le {\rm min}(J_{r+1}, j+m-1)$.\\ 
			For each pair $(j,k)\in {\cal P}_{r+1}$: let $S_{j,k}^{r+1}$ be the section obtained by merging sections $S_j^{r+1}, \dots, S_k^{r+1}$ and compute the histogram resulting from this merge:
			$h_{j,k}^{r+1} = \sum_{i=j}^k h_i^{r}(c),~c\in {\cal C}$, and compute the values of ${\hat {\cal L}}(S_{j,k}^{r+1})$ and the the merging gain $g_{j,k}^{r+1} = \sum_{i=j}^k l_j^{r+1} - {\hat {\cal L}}(S_{j,k}^{r+1})$ using the histogram and Eq.~\eqref{Eq:Ls}. In most cases the values will be copied directly from corresponding values of $h_{j,k}^r(c), ~c \in {\cal C}$, ${\hat {\cal L}}(S^r_{j,k})$ and $g_{j,k}^r$. Actual calculation is needed only if $j<j^*\le k<k^*$, that is for $m(m+1)/2$ entries.
			\ENDFOR	
			\STATE 3. Let $r^*$ be the final value of $r$, $J^*= J_{r^*}$ be the final number of sections and $S^*_j = S_j^{r^*}$, $h_j^*(c) = h_j^{r^*}(c)$, $j=1,\dots, r^*, c\in {\cal C}$ be the final sections and corresponding histograms.		
		\end{algorithmic}
		\label{alog:Coding}
\end{algorithm*}	} }

\subsubsection{Partitioning into AC Sections} 
The partition of the codeword sequence into sections is preformed using a greedy algorithm (Algorithm~\ref{alog:Coding}). We begin by partitioning the sequence into RLE sections, and in each iteration we merge up to $m$ consecutive sections, so that the merging yields the greatest possible reduction in total coded sections length; $m$ is a constant which was set to 4 in our experiments. When the algorithm starts, all the bits are spent on HFSs and histogram representation, and none on AC bit sequences. As the algorithm progresses and sections are merged, more bits are spent on AC, and the histograms become fewer in number, but having more non-zero counts.

{Determining the effect of merging several consecutive AC sections requires repeated calculation of the the coded length of AC sections. In principle this could be done by performing arithmetic coding and checking the resulting length. However, doing that in each iteration for each potential sections merge would be prohibitively computationally intensive. Instead we use the tight lower bound of \eqref{Eq:H_codeword} as an approximation to the section length.
Let  $\{h_s(c), c\in {\cal C}\}$ be the histogram of section $s$, where $h_s(c)$ is the count for codeword $c$, 
and let }
\begin{equation}
{\textstyle M_s = \sum_{c\in {\cal C}} h_s(c)} \nonumber
\end{equation}
{be the number of codewords in the merged section. The codeword probability is approximated by $ p_c \approx h_{s}(c) / M_s,  c \in {\cal C} $.} Hence, using \eqref{Eq:H_codeword} and rounding up the result to the nearest multiple of 8 (for byte alignment), the estimated AC coded length of section $s$ is
\begin{equation}
{\textstyle {\hat{\cal L}_{\rm AC}(s) \stackrel{\rm def}{=}8 \left\lceil \sum_{c\in{\cal C}} h_s(c) \log_2(M_s/h_s(c))/8 \right\rceil}}, \label{Eq:Lac}
\end{equation} 
{where $\lceil a \rceil$ is the least integer not smaller than $a$.}
{The total section code length, including the histogram is estimated by
\begin{equation}
{\textstyle \hat{\cal L}(s) = \hat{\cal L}_{\rm AC}(s) + {\cal L}_{\rm H}(s) +2}, \label{Eq:Ls}
\end{equation}
where ${\cal L}_H(s)$ is the number of bits used for the histogram coding and the last term on the right hand side is the bits used for the HFS.}


\subsubsection{Complexity of Algorithm~\ref{alog:Coding}}
To analyze the amount of computations, we note that $J_r - m \le J_{r+1}<J_r$ for $0\le r <r^*$ and therefore $J_r \le M-r$ and $r^* \le M - J_{r^*} \le M-1$. We also note that
$|{\cal P}_r| = \sum_{i=1}^m (J_r -i) = m(J_r -(m+1)/2)\le m(M-r - (m+1)/2)$.

Consider the steps of the algorithm: 
\begin{itemize}
	\item Steps 1a and 1b are of complexity $O(M)$;
	\item Step 1c is of complexity $O(1)$;
	\item Step 1d is of complexity $O(|{\cal P}_0|) = O(m(M-(m+1)))$.
\end{itemize}
Thus the initialization complexity is $O((m+1)M-m)$.

In step 2a, finding the maximum among $J_r$ elements requires $O(\log_2(J_r)) \le O(\log_2(M-r))$ operations, thus in total step 2a requires (using Stirling's formula) 
\begin{align}
&{\textstyle\sum_{r=0}^{r^*-1}O(\log_2(J_r))} \le {\textstyle\sum_{r=0}^{M-1} O(\log_2(M-r))} \nonumber\\
&\qquad = {\textstyle O(\log_2 M!) = O(M(\log M-1))}.
\end{align}
Assuming that the arrays are organized in a tree structure, which allows insertion and deletion of elements at a fixed complexity, each iteration of step 2b and 2c is of complexity $O(1)$  and the complexity of step 2d is $O(m(m+1)/2)$.  Since the iterations are repeated at most $(M-1)$ times, the expression gives the complexity of Algorithm 1 as
\begin{equation}
{\textstyle O(M(\log M + m(m+1)/2))}.
\end{equation}
Since $M\ll N$ and the complexity of computing the measurements is $O(N\log N)$, assuming a sensing matrix based on a fast transform, the complexity of performing the entropy coding is significantly lower than that of computing the measurements.

{\textbf {A note about the significance of choice of  $m$}}: This is a greedy algorithm which stops when it cannot make an improvement by merging up to $m$  adjacent sections. Increasing $m$  will make the algorithm consider more options in each iteration, and therefore potentially find a better solution. Thus it appears that choosing $m$  is a trade-off between complexity and better compression. However, if the algorithm chooses to merge more than two sections at a time it skips an iteration, therefore choosing a larger $m$  may sometimes improve performance. 
In our simulations we used $m=4$.
%
\textbf{\begin{figure}[htbp!]
		\begin{center}
			\includegraphics[width=.9\linewidth]{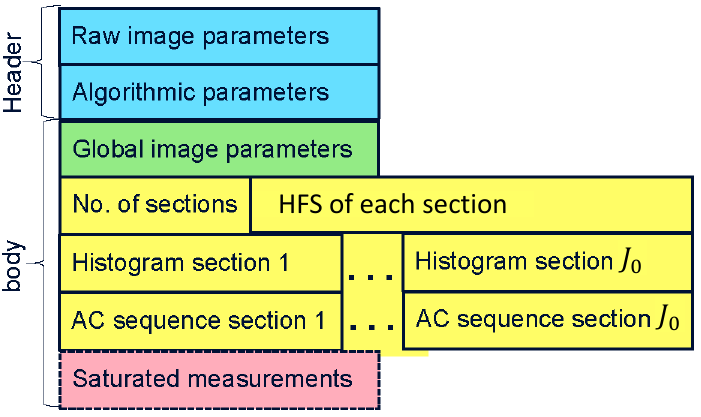}
		\end{center}
		\vspace{-2mm}
		\caption{Coded image structure.}
		\label{fig:structure}
\end{figure}}

\subsubsection{Coded Image Structure}
The data structure of the coded image is shown in Fig.~\ref{fig:structure}. The coded image begins with a header specifying the image parameters (size, bits per pixel, color  scheme, \etc) and the encoder's algorithmic choices (\eg, sensing matrix type, compression ratio, quantizer parameters). The body of the code consists of some global image parameters ($\mu$, $s$, $L$, $\tilde{Q}(y_0)$, \etc) followed by the entropy coded quantized measurements, which are the bulk of the transmitted data: the HFSs, the histograms and the AC sequences for each section. If the values of $\tilde{Q}(y)$ are transmitted for saturated measurements (corresponding to terms equaling $L$ in the codeword sequence), these are coded as an array of unbounded signed integers.

\subsection{Decoding}
Decoding is done in reverse order of encoding, as follows:
\begin{itemize}
\item
	The numerical parameters and bit arrays in the coded image are parsed.
\item
	The quantization codewords are recovered by arithmetic decoding.
\item
	The unsaturated quantized measurements are computed using \eqref{Eq:Dequant}. For the saturated measurements (those having a codeword of $L$), if values of $\tilde{Q}(y)$ are transmitted, they are used. Otherwise, the values of the quantized saturated measurements are set to zero.
\item
	The sensing matrix is determined according to the algorithmic choices in the coded image. If there was no \adhoc transmission of $\tilde{Q}(y)$ for saturated measurements, the rows corresponding to these measurements are set to zero. In practice this is done by replacing the original sensing matrix $\Phimat$ by $\Dmat \Phimat$, where $\Dmat$ is a $M \times M$ diagonal matrix whose diagonal elements are zero for saturated measurements and one for unsaturated measurements.
\item
	The image is reconstructed using the sensing matrix and the quantized measurements, as described in detail in Sec.~\ref{Sec:rec}.
\end{itemize}

\subsection{Module Comparison of CSbIC  with JPEG and JPEG2000 \label{Sec:CompareJPEG}}
\begin{figure*}[htbp!]
	\centering
	\includegraphics[width=1.0\linewidth]{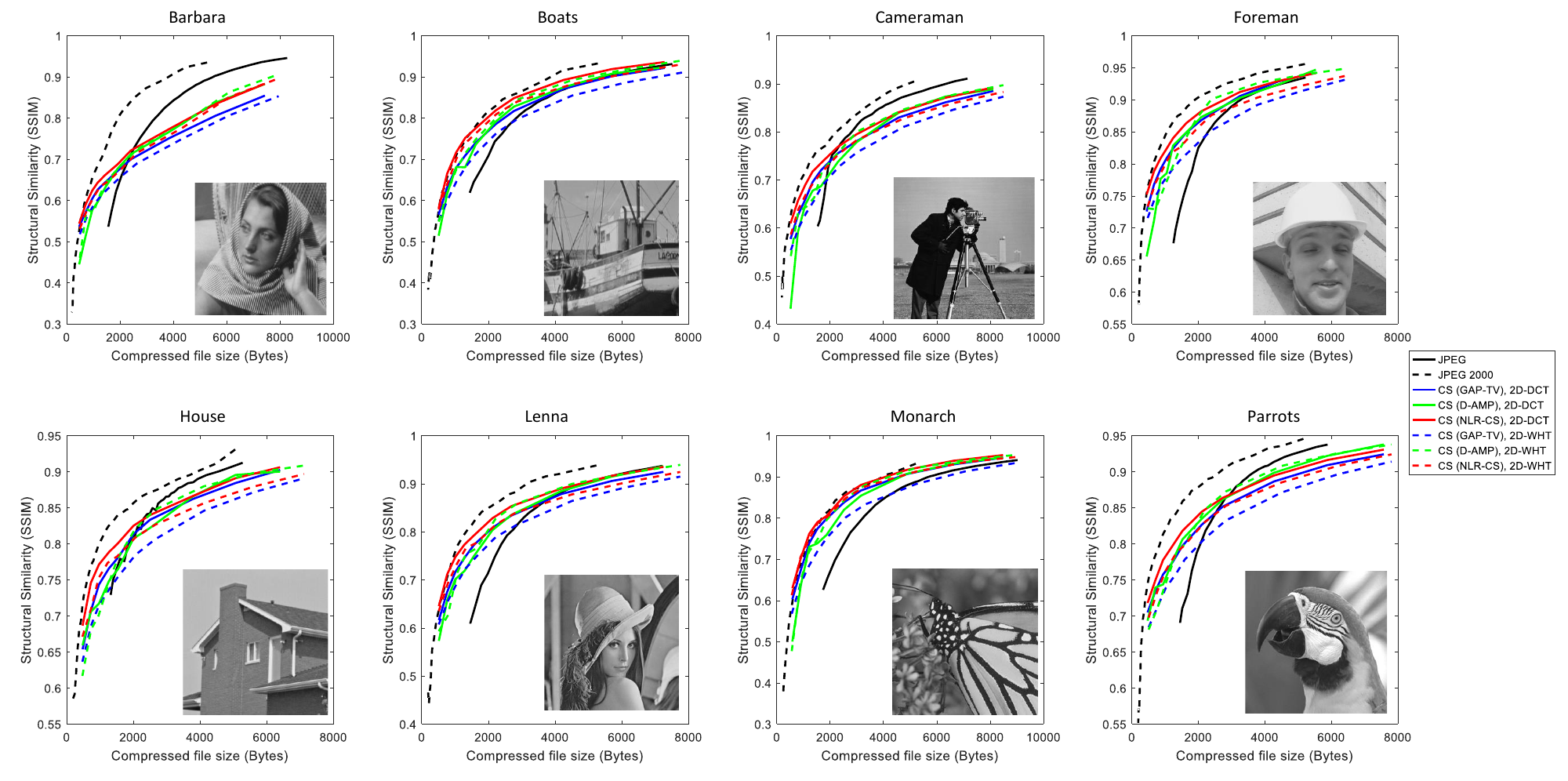}
	\vspace{-4mm}
	\caption{Performance diagrams, SSIM vs. compressed file size (in bytes), comparing JPEG  (black solid curves), JPEG2000 (black dash curves) with CSbIC compression using different sensing matrices -- 2D-DCT (solid) and 2D-WHT (dash), and different reconstruction algorithms --- GAP-TV (blue), NLR-CS (red) and D-AMP (green).}
	\label{fig:SSIM_8img_3algo}
	\vspace{-5mm}
\end{figure*}

We now compare the architectures of CSbIC, JPEG and JPEG2000 and consider the aspects that may lead to performance differences. Fig.~\ref{Fig:Arch} compares the architecture of CSbIC with that of JPEG side by side. The main common points and differences are: 

\begin{itemize}
	\item
	JPEG, JPEG2000 and CSbIC begin with a linear projection of the image onto a different space. However:
\begin{itemize}
		\item[--]
		JPEG2000 may partition the image into tiles of varying sizes, which are processed separately. This is equivalent to using a block-diagonal projection matrix, where each block corresponds to a tile. In JPEG the tiles (referred to as blocks) are of fixed size of  $8 \times 8$ pixels. In CSbIC the projection is done on the whole image, which was adequate for the image sizes that we experimented with. 
		\item[--]
		In both JPEG and JPEG2000, each tile/block is projected on a space of the same dimension, $N$, hence there is no data loss in this operation, whereas in CSbIC the projection is lossy since it is on a $M$-dimensional space, $M \ll N$.
		\item[--]
		In JPEG the projection is a 2D-DCT with the output organized in zig-zag order, while in JPEG2000 it is a 2D-wavelet transform. In CSbIC, a 2D-DCT based projection is one of several options.
		\item[--]
		JPEG uses block to block prediction of the DC coefficient in order to deal with the issue of the DC coefficient being much larger than the other coefficients. In JPEG2000 this is done by subtracting a fixed value from all pixels before the projection. In contrast, CSbIC takes care of this issue in the quantization stage. The effect of these different methods is similar and has little impact on performance. 
	\end{itemize}
	\item
	CSbIC uses a simple uniform scalar quantizer with an identical step size for all measurements. JPEG uses the same type of quantizer, but the step size is selected from a quantization table and is different for each coefficient, resulting in quantization noise shaping, which may give JPEG an advantage at higher quality/data rate (Fig.~\ref{fig:SSIM_8img_3algo}). JPEG2000 also performs quantization noise shaping through varying step size, and in addition, its quantizer is not exactly uniform---the quantization interval around zero (the ``dead-zone'') is larger than the other quantization intervals, effectively forcing small wavelet coefficients to zero and reducing the amount of bits spent on coding them. 
	\item
	The bit rate and quality trade-off in JPEG and JPEG2000 is controlled by tuning the operation of a single module --- the quantizer. In contrast, in CSbIC this trade-off is controlled by jointly tuning two different modules: The projection, or measurement capturing module is tuned by changing the compression ratio, and the quantizer is tuned by changing the quantization step.
	\item
	In JPEG, entropy coding is based on Huffman coding and RLE. JPEG2000 uses arithmetic coding, with a sophisticated adaptive algorithm to determine the associated probabilities. CSbIC uses arithmetic coding, which is known to be better than Huffman coding, but instead of using adaptive estimation of the probabilities, the codewords are partitioned into sections and for each section a histograms of codewords is computed and sent as side information. The overhead of the transmitted histograms may be a disadvantage of CSbIC relative to JPEG2000.
	\item
	In JPEG and JPEG2000, the decoder generates the image from the dequantized transform coefficients by an inverse 2D-DCT or wavelet transform, respectively --- a simple linear operation that does not rely on any prior information not included in the received data. In contrast, the CS reconstruction in CSbIC is an iterative, non-linear optimization, which relies on prior assumptions about the structure of the image (\eg, sparsity) or some recently developed deep learning based algorithms.
\end{itemize}

\section{Image Reconstruction \label{Sec:rec}}
Since the seminal works of Cand\`{e}s \etal\cite{Candes06ITT} and Donoho~\cite{Donoho06ITT}, various reconstruction algorithms have been developed.
The early reconstruction algorithms leveraged the property of natural images of being compressible when projected by a suitable sparsity operator $\Dmat$:
\begin{equation}
\fv = \Dmat\xv ,
\end{equation}
where $\fv$ denotes the projected vector and it is usually forced to be sparse. 
$\Dmat$ can be a pre-defined basis (DCT or wavelet), or learned on the fly~\cite{Aharon06TSP}. Another popular sparsity operator is the Total Variation (TV). 
%

Recently, better results were obtained by algorithms such as D-AMP (denoising based approximate message passing)~\cite{Mertzler14Denoising} and NLR-CS (nonlocal low-rank compressive sensing)~\cite{Dong14TIP}, which exploit established image denoising methods or the natural images property of having a low rank on small patches. While using different projection operators $\Dmat$, most of these algorithms compute $\hat{\xv}$, the estimated signal, by solving the minimization problem
\begin{equation}
{\textstyle \hat{\xv} = \argmin_{\xv} \|\Dmat \xv\|_p, ~~{\rm s.t.}~~ \yv = \Phimat \xv}, \label{Eq:MinSprs}
\end{equation}
where $p$ can be 0 ($\|\fv\|_0$ denotes the number of non-zero components in $\fv$) or 1, using $\|\fv\|_1$ as a computationally-tractable approximation to $\|\fv\|_0$. Alternatively, $\|~\|_p$~can stand for the nuclear norm $\|~\|_*$ to impose a low rank assumption~\cite{Dong14TIP,Yuan15GMM,Zhang18CVPR,Zha2020RRC_TIP}. 

Problem~\eqref{Eq:MinSprs} is usually solved iteratively, where each iteration consists of two steps.
Beginning with an initial guess, the first step in each iteration projects the current guess onto the subspace of images, which are consistent with the measurements, and the second step denoises the results obtained in the first step~\cite{Yuan18SP}. 
%
The denoising based image compressive sensing reconstruction algorithm (D-AMP), proposed in~\cite{Mertzler14Denoising}, integrates various denoising algorithms~\cite{Dabov07BM3D} into the approximate message passing (AMP)~\cite{Donoho09AMP} framework and leads to the state-of-the-art reconstruction results using Gaussian sensing matrices. AMP extends iterative soft-thresholding by adding an extra
term to the residual known as the Onsager correction term.
%
By exploiting the nonlocal self similarity of image patches and low-rank property, Dong {\em et al.}~\cite{Dong14TIP} proposed the nonlocal low-rank regularization based compressive sensing reconstruction algorithm, \ie, NLR-CS. 
NLR-CS is an iterative optimization algorithm based on the alternating direction method of multipliers (ADMM)~\cite{Boyd11ADMM} framework. 
In each iteration, after the updating of $\xv$, instead of using various denoising algorithms as in D-AMP, NLR-CS imposes low rank property of similar patches to achieve a better image.\footnote{Some other algorithms~\cite{Zhang14TIP,Shi17ICME} has also been proposed based on the group sparsity for CS problem, but most of them are based on the block-CS sampling, which is different from the problem considered here.}

In our experiments we have used an improved variant of TV known as Generalized Alternating Projection Total Variation (GAP-TV)~\cite{Yuan16ICIP_GAP}, as well as D-AMP and NLR-CS. 

\section{Performance Testing \label{Sec:perf_mea}}
\begin{figure}[htbp!]
	\centering
	\vspace{-1mm}
	\includegraphics[width=\linewidth,height = 4.5cm]{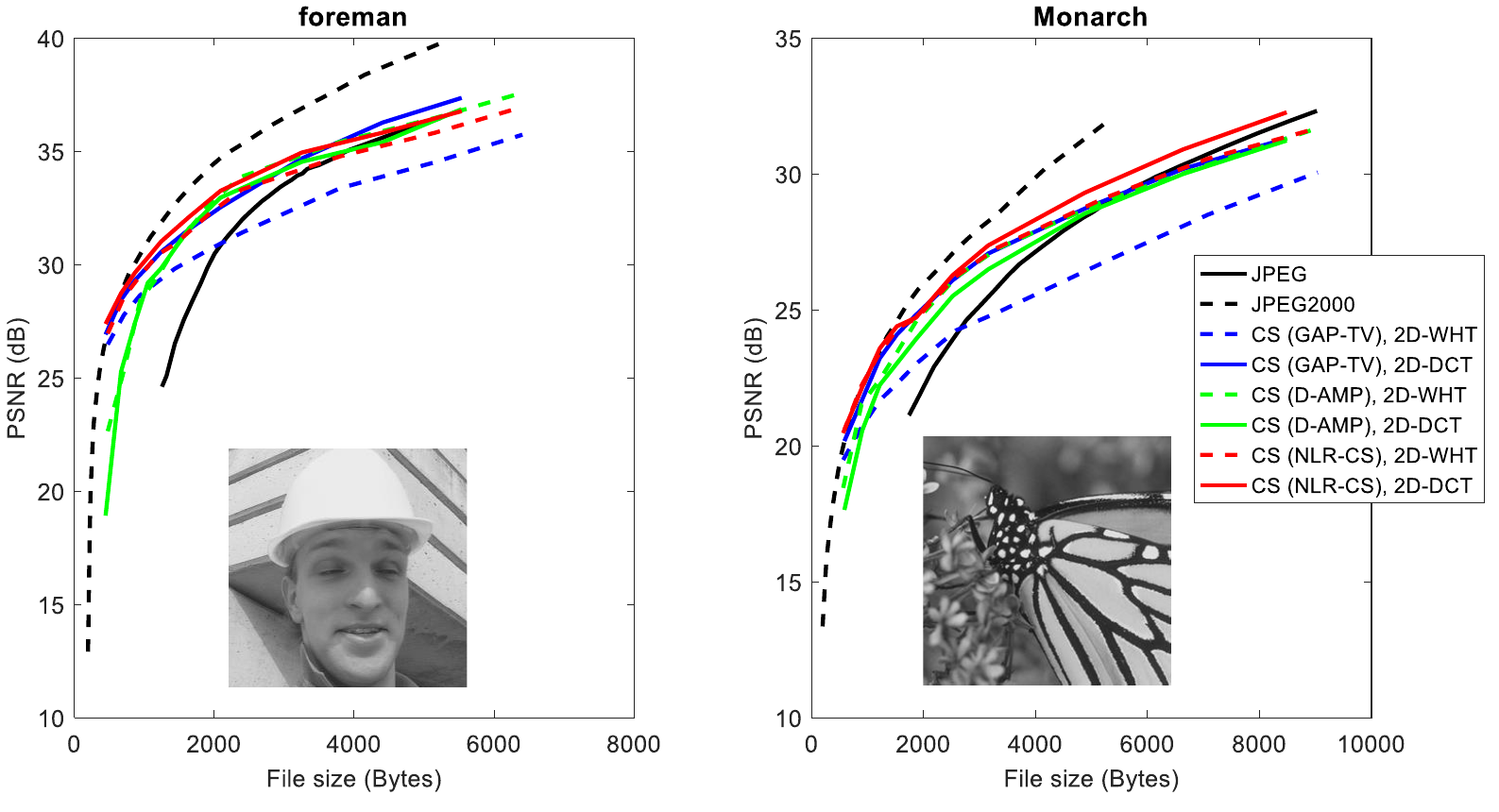}
	\vspace{-5mm}
	\caption{PSNR vs. compressed file size (in bytes), comparing JPEG (black
		solid curves), JPEG2000 (black dash curves) with CSbIC compression using
		different sensing matrices – 2D-DCT (solid) and 2D-WHT (dash), and
		different reconstruction algorithms — GAP-TV (blue), NLR-CS (red) and
		D-AMP (green). PSNR results for all 8 images are available in~\cite{CSvsJPEG_webpage}.}
	\label{fig:PSNR_8img_3algo}
\end{figure}

Our initial test material consisted of 8 widely used monochrome images of 256x256 8-bit pixels (Fig.~\ref{fig:SSIM_8img_3algo}). These images were processed in a variety of test conditions, specified by encoder and reconstruction parameters. The outcome of processing an image in a particular test condition is the data rate, as expressed by the size of the coded image file (in bytes), and the reconstructed image quality, measured by structural similarity (SSIM)~\cite{Wang04imagequality}. We preferred SSIM over Peak-Signal-to-Noise-Ratio (PSNR) as a measure of quality because SSIM better matches the quality perception of human visual system (refer to Fig.~\ref{fig:PSNR_8img_3algo} and~\cite{CSvsJPEG_webpage} for PSNR results, which provide similar observations to SSIM).

\begin{figure}[htbp!]
	\centering
	\includegraphics[width=1\linewidth]{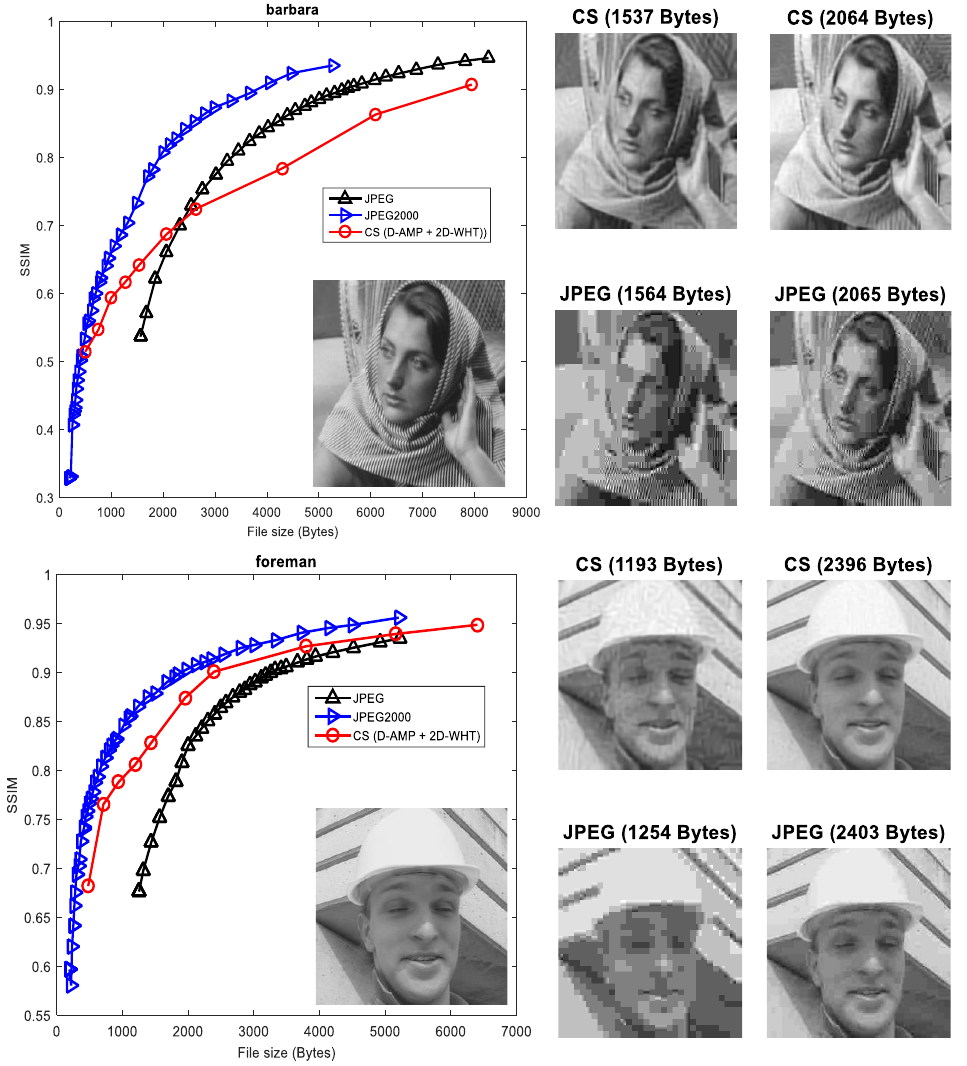}
	\vspace{-5mm}
	\caption{Left: SSIM vs. compressed files size (in bytes) D-AMP+2D-WHT is used in CSbIC. Right: exemplar reconstructed images from CSbIC and JPEG with similar file size. More results are available at~\cite{CSvsJPEG_webpage}.}
	\label{fig:imges_cs_JPEG}
	\vspace{-1mm}
\end{figure}

\begin{figure}[htbp!]
	\begin{center}
		\includegraphics[width=.5\textwidth]{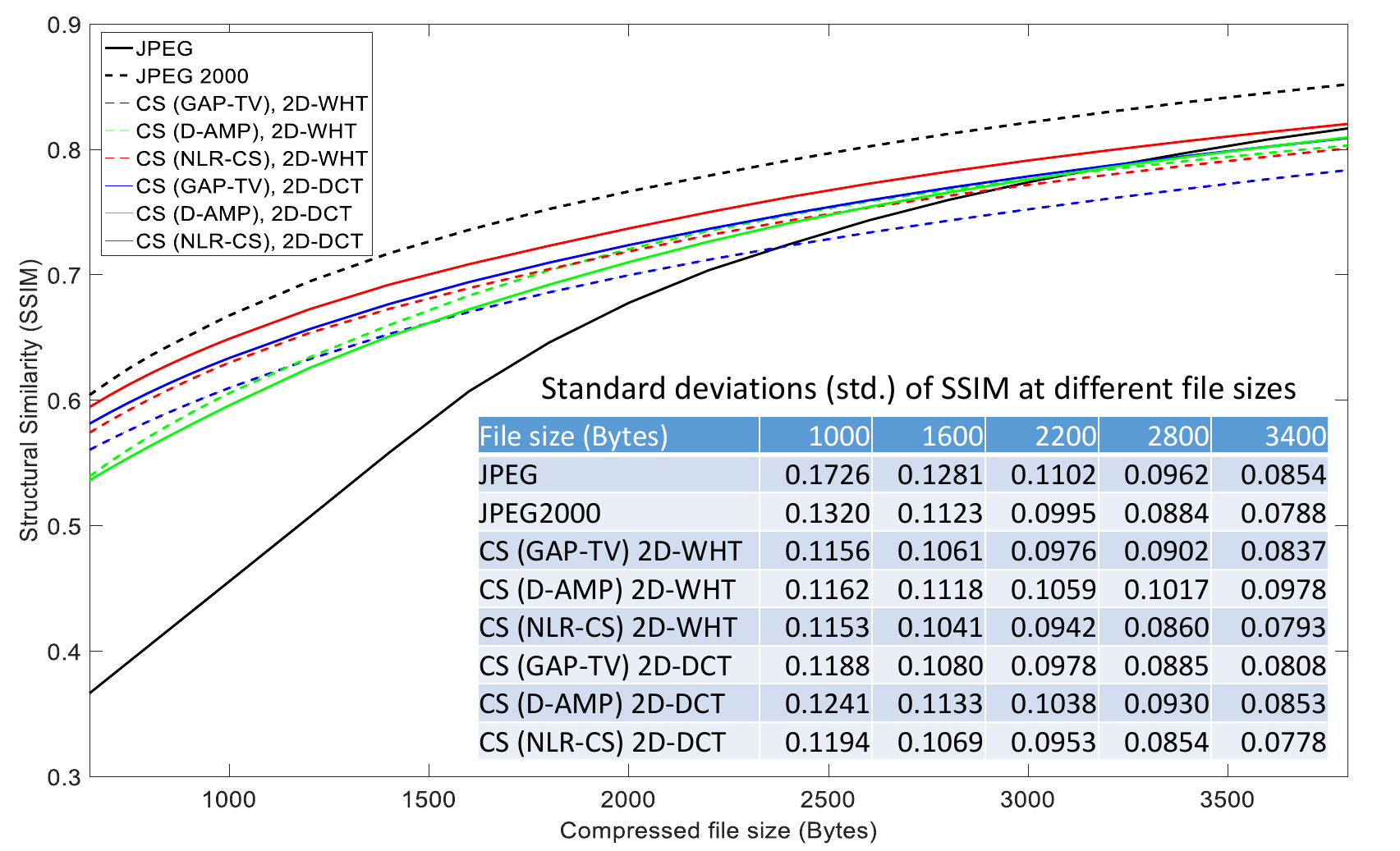}
	\end{center}
	\vspace{-0.3cm}
	\caption{Average SSIM vs. compressed file size (in bytes), computed over 200 images from BSDS300 dataset, comparing JPEG (black solid curves), JPEG2000 (black dash curves) with CSbIC compression using 
		different sensing matrices. The SSIM standard deviations at various compressed 
		sizes are shown in the table. Below 4000 bytes NLR-CS/2D-DCT outperforms JPEG on average and has lower variances, which shows better robustness to variations among images.
		Exemplary results for individual images are available at~\cite{CSvsJPEG_webpage}. }
	\label{Fig:8img_BSDS}
\end{figure}

Fig.~\ref{fig:imges_cs_JPEG} shows a few examples where higher SSIM is clearly consistent with higher image quality.  
Results are presented as points in SSIM vs. coded image size diagram. By connecting the points corresponding to test cases with different rate control parameters (but identical in all other parameters) we get performance curves which can be compared with each other to determine the better operating parameters. Performance curves for JPEG or JPEG2000 were obtained using the MATLAB implementation of those standards (the ``imwrite" function).
We confirmed our results by repeating our tests on 200 images from the BSDS~300 dataset~\cite{MartinFTM01}. 
Fig.~\ref{Fig:8img_BSDS} shows SSIM average and standard deviation, over the 200 images, vs. compressed file size.
The 2D-DCT sensing matrix with NLR-CS reconstruction performs better than JPEG for compressed file size less than 4000 bytes.  
It is also more robust to variations among images, as shown by its significantly lower standard deviation.
All the results presented in this section are supported by performance diagrams of additional images in~\cite{CSvsJPEG_webpage}, along with the source code.

\subsection{Effect of The Choice of Sensing Matrix \label{Sec:choice_sensing}}
%

\begin{figure}
	\centering
	\vspace{-1mm}
	\includegraphics[width=1\linewidth]{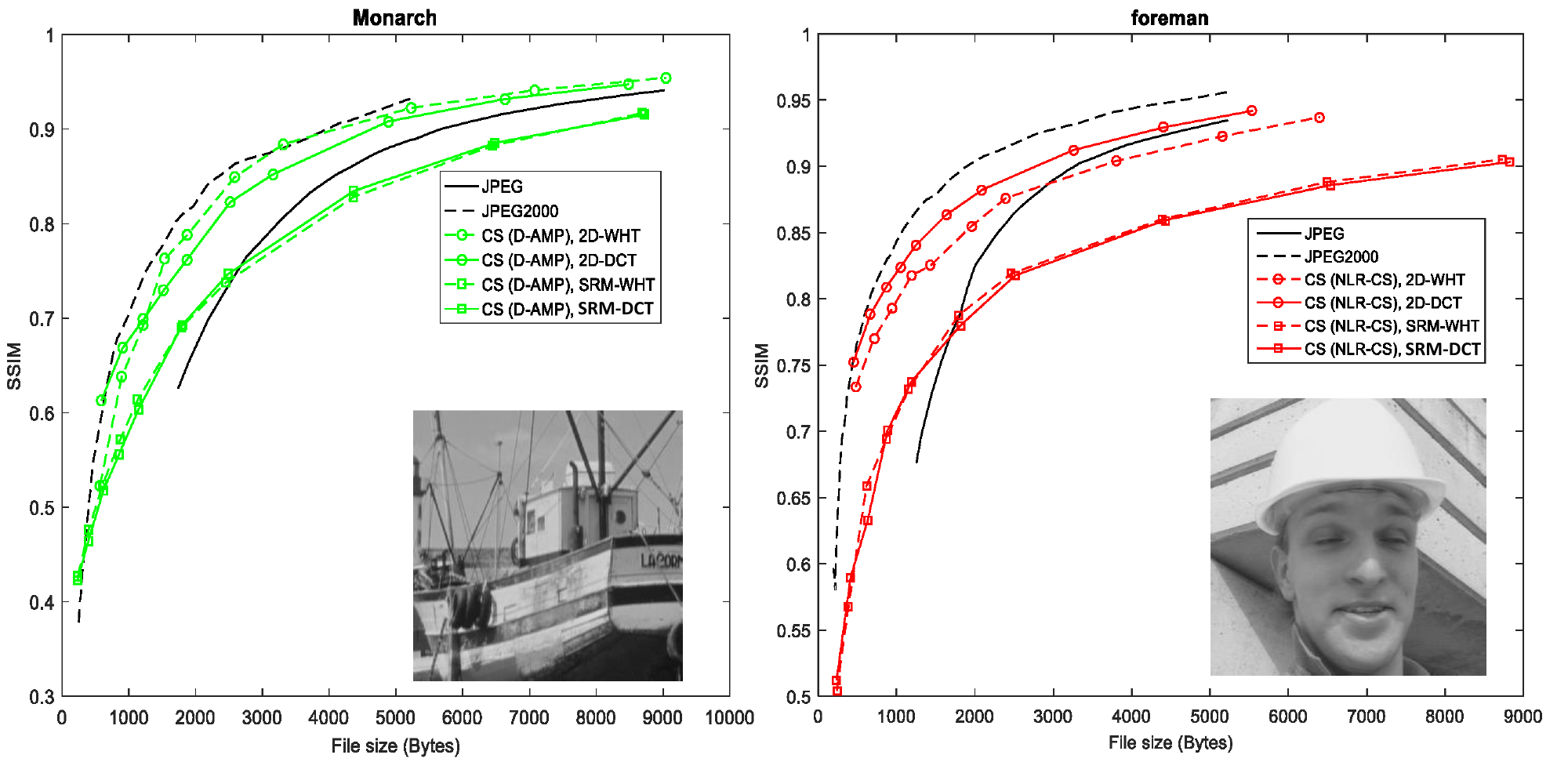}
	\vspace{-3mm}
	\caption{Performance plots for the two images of CS with different sensing matrices. D-AMP (left) and NLR-CS (right) are used for reconstruction. Results with 8 images and 3 algorithms are available at~\cite{CSvsJPEG_webpage}.}
	\label{fig:diff_sensM}
\end{figure}

The performance with the deterministic sensing matrices, 2D-DCT and 2D-WHT, was always significantly better than with the SRMs with the same fast transforms, SRM-DCT and SRM-WHT, respectively (Fig.~\ref{fig:diff_sensM}), regardless of the reconstruction algorithm that was used. Within each of those groups (deterministic matrices and SRMs), the sensing matrices based on DCT generally yielded better performance than the those based on WHT (Fig.~\ref{fig:diff_sensM} right). However, (i) the difference, in terms of SSIM for the same file size, is generally smaller than the performance difference between deterministic and structurally random matrices, (ii) the difference between SRM-DCT and SRM-WHT is generally much smaller than the difference between 2D-DCT and 2D-WHT, and (iii) the magnitude of the difference varies with the image and the reconstruction algorithm that is used. In particular, with GAP-TV the performance difference between DCT and WHT based matrices was most pronounced, whereas with D-AMP WHT performance was very close to that of DCT, and in a few cases the WHT based matrices slightly outperformed the DCT based ones. 

\begin{figure}[htbp!]
	\begin{center}
		\includegraphics[width=.5\textwidth]{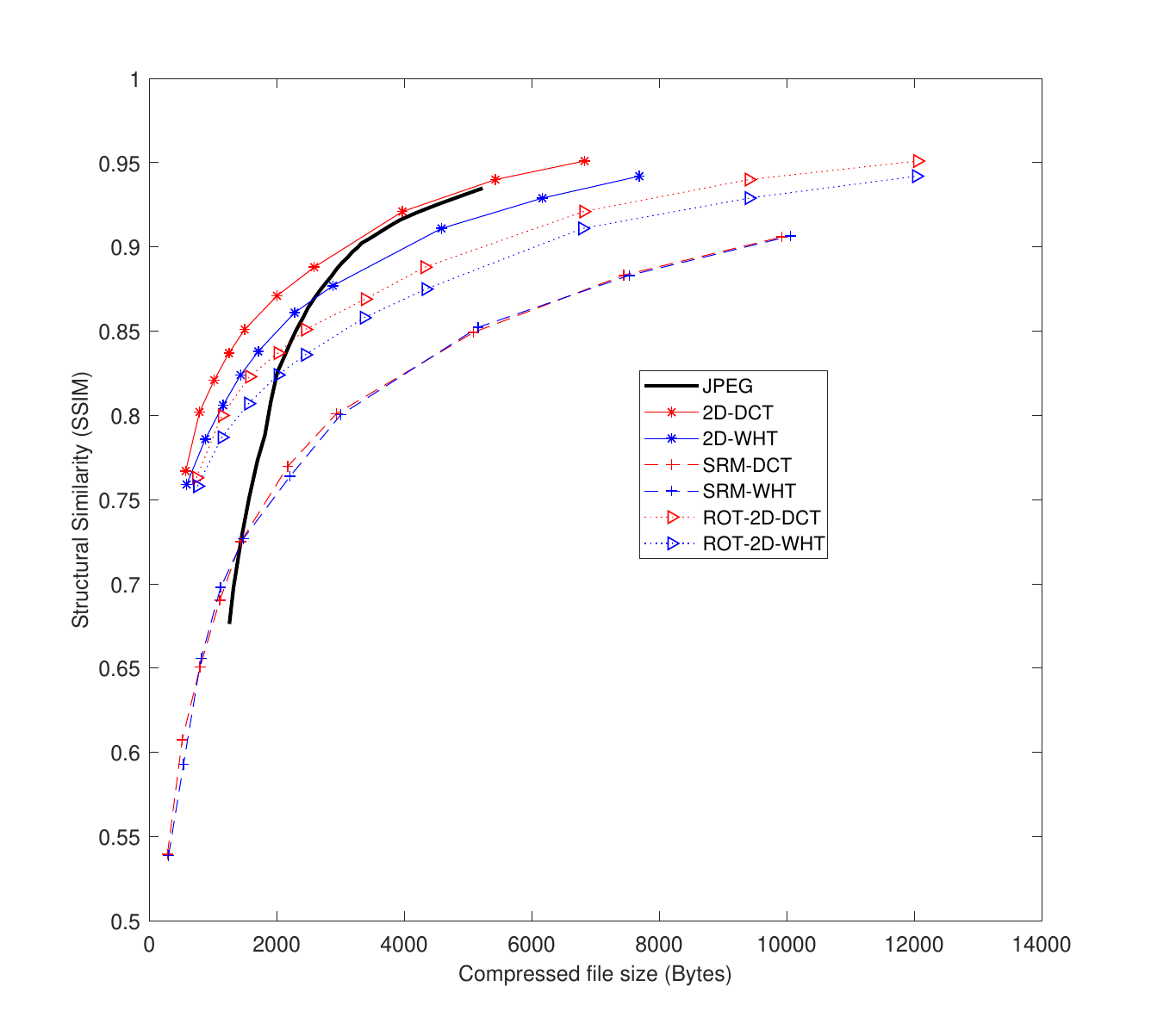}
	\end{center}
	\vspace{-0.2cm}
	\caption{Reconstructed image quality vs. file size of ``Foreman" image with deterministic (2D-DCT and 2D-WHT), SRM and ROT sensing matrices.}
	\label{Fig:Foreman_ROT_2D_SRM}
\end{figure}
An obvious reason for these differences is the fraction of the signal energy that is captured by the measurements with each sensing matrix. The DC measurement is the same, up to a scaling factor, in all four matrices. Therefore, when speaking about signal energy and measurements we refer only to the AC component and exclude the  DC component. Since SRMs randomize and whiten the signal prior to applying the transform, their measurements capture about $M/N$ of the signal energy. The 2D-DCT measurements capture a much larger fraction of the signal energy because most of the signal energy is concentrated in the low frequencies. Since 2D-WHT is a crude approximation of spectral decomposition, its low ``frequency" components capture less energy than those of 2D-DCT, but much more than the $M/N$ of the SRM matrices. This argument is certainly true, but it is only a part of the explanation. Consider the sensing matrix given by
\begin{equation}
\tilde{\Phimat} = \Thetamat \Phimat, \label{Eq:PrjctdSnsMtrx}
\end{equation}
where $\Phimat  $ is a $M \times N$  2D-DCT or 2D-WHT deterministic sensing matrix and  $\Thetamat$ is given by
\begin{equation}
\Thetamat = \left[\begin{array}{cc} {\bf 1} & {\bf 0}_{1\times(M-1)} 
\\ {\bf 0}_{(M-1)\times 1} &  \tilde{\Thetamat} \end{array} \right], \label{Eq:ThetaDef}
\end{equation}
where $\tilde{\Thetamat}$ is a random orthonormal $(M-1) \times (M-1)$ matrix, defined by
\begin{equation}
\tilde{\Thetamat} = {\bf SHR},  \label{Eq:tildeThetaDef}
\end{equation}
where $\Rmat$ is a diagonal matrix whose diagonal entries are IID random variables that have the values of $\pm 1$ with equal probability; $\Hmat$ is the DCT transform matrix of  order $M-1$, with rows scaled to have unit norm; and $\Smat$ is a random selection matrix, \ie, a matrix in which each row and each column contains one entry of 1 and the rest are zero.

\begin{figure}[h]
	\centering
	\includegraphics[width=.5\textwidth]{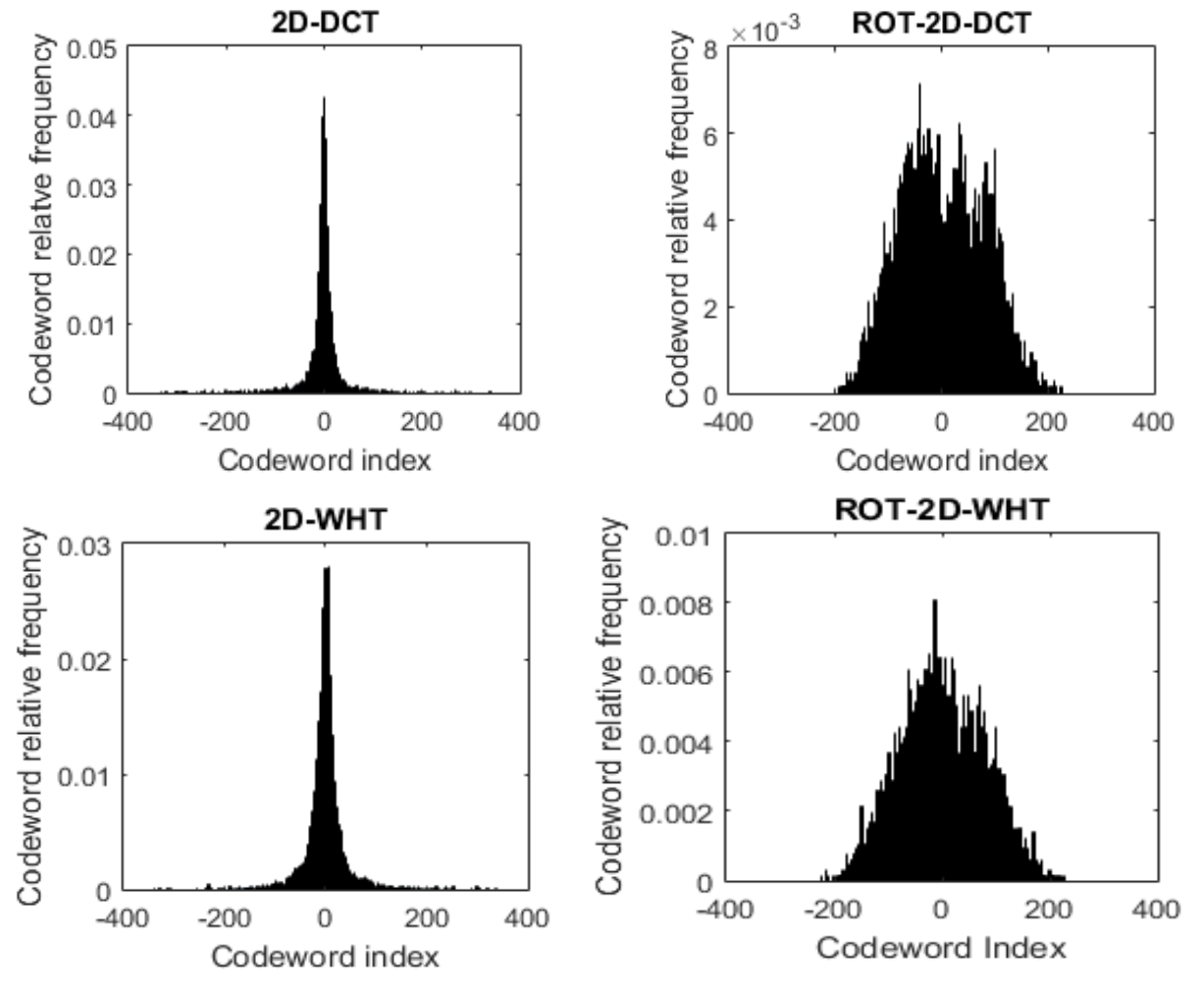}
	\vspace{-0.5cm}
	\caption{Codeword histograms of of ``Foreman" for deterministic and ROT sensing matrices and with compression ratio of 0.1. The distribution of the codewords in the deterministic case is much tighter than for the ROT matrices, and it is tighter for 2D-DCT.}
	\label{Fig:Foreman_codeword_hist}
\end{figure}
The measurements computed by $\tilde{\Phimat}$ are obtained by a random orthonormal transformation (ROT) of the measurements computed by $\Phimat$, hence we denote the sensing matrices $\tilde{\Phimat}$  ROT-2D-DCT and ROT-2D-WHT, respectively. The ROT matrices capture the same components of the signal as the corresponding deterministic matrices and the addition of the orthonormal transformation $\Thetamat$ should not matter to any of the reconstruction algorithms. Furthermore, the quantization step, given by Eq.~\eqref{Eq:RSN}, is the same and so are the quantization noise and digitization noise, given by Eq. \eqref{Eq:DigitizationSigma}. Nevertheless, the performance of the ROT matrices, while significantly better than the SRM matrices, was not as good as the deterministic ones. The difference in SSIM between the performance curves of a deterministic matrix and the corresponding ROT matrix was about a third or a quarter of the difference between the performance curves of the same deterministic matrix and the corresponding SRM matrix (see example in Fig.~\ref{Fig:Foreman_ROT_2D_SRM}). Thus, multiplying the measurements vector $\Phimat \xv$ by $\Thetamat$, causes a significant performance degradation. By its definition, $\tilde{\Thetamat}$ is a DCT-based SRM with a compression ratio of 1, which uses \emph{local randomization}, that is, the input signal is randomized by randomly toggling the signs of its entries. This randomization method is different from the \emph{global randomization} used in the SRM-DCT sensing matrix, where the input signal is randomized by randomly permuting its entries, but both variants have a similar effect on the distribution of the measurements: when a vector is multiplied by either one of these matrices, the distribution of the entries of the result is approximately zero mean Gaussian or a mixture of zero mean Gaussians with similar covariances \cite{Haimi-CohenL16_SP}. 

Figure~\ref{Fig:Foreman_codeword_hist} shows the effect of multiplication by  $\tilde{\Thetamat}$ on the AC measurements. The measurements generated by the deterministic matrices have a narrowly peaked distribution with a long tail, while the measurements generated by the ROT matrices have a much wider, Gaussian like distribution, with a shorter tail. As a result, the entropy coding needs significantly more bits per measurement for coding the ROT measurements. The distribution of the measurements generated by SRM-DCT and SRM-WHT has also a Gaussian-like shape, similar to that of the ROT generated measurements. Therefore, we may conclude that the difference between the SRM and the corresponding deterministic matrices has two components: one component is the different fraction of signal energy captured by each method, and is represented by the improvement from the SRM matrices performance to that of the ROT matrices. A second component relates to the different distributions of the measurements in the deterministic case and it is represented by the improvement from the ROT matrices performance to that of the corresponding deterministic matrices.

One can notice in Figure~\ref{Fig:Foreman_codeword_hist} that the histogram of 2D-WHT is a little wider than that of 2D-DCT. Thus, the advantage of 2D-DCT over 2D-WHT seems also to be not only because of the amount of signal energy captured in each case but also because of differences in the measurements distribution. 
For all the three algorithms, SRM-DCT performs similar to SRM-WHT, since the captured fraction of signal energy, and the measurements distribution are similar in both cases. 
   
%
%
%
%
%
%
\begin{figure}[tbp!]
	\centering
	\includegraphics[width=.5\textwidth]{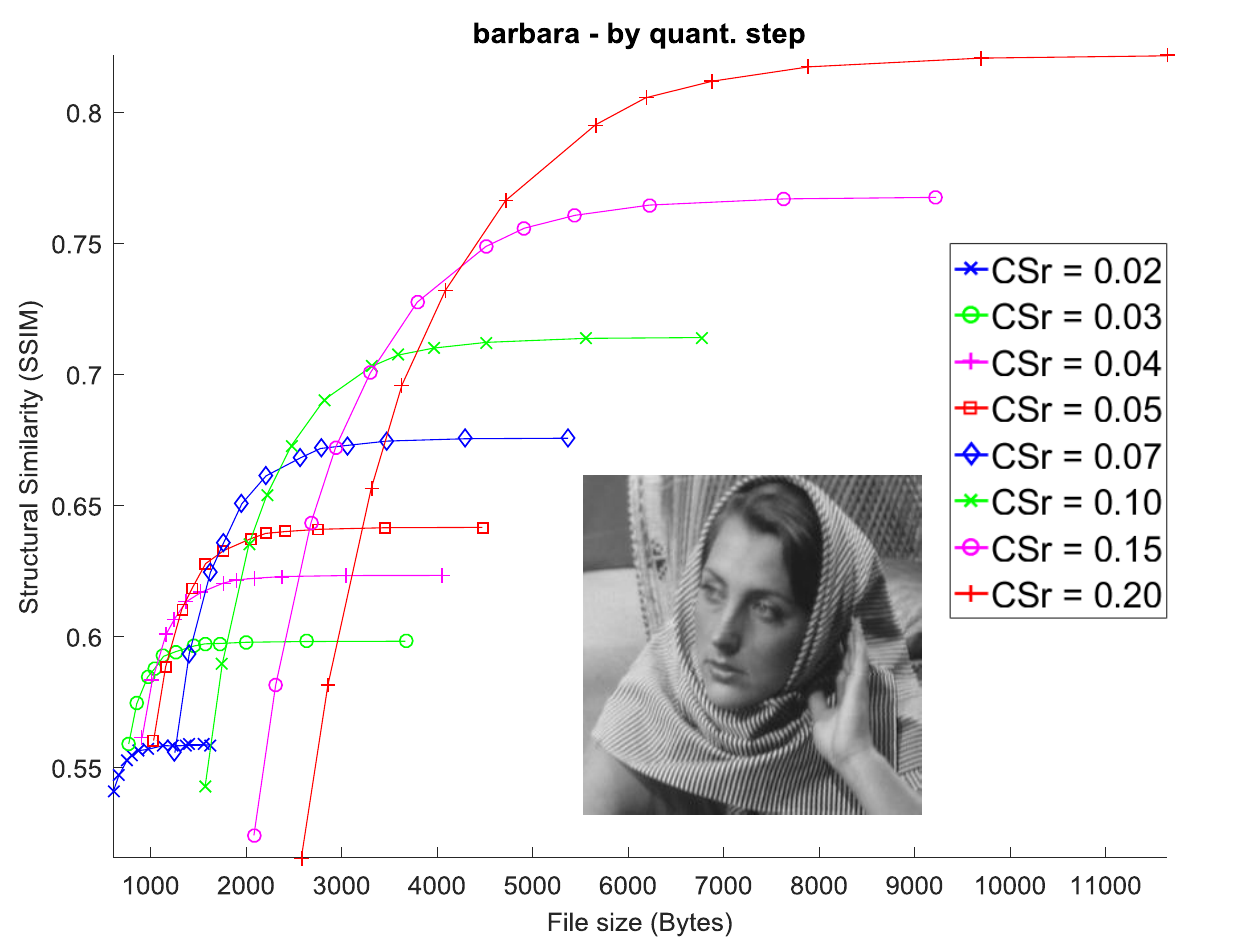}
		\vspace{-0.5cm}
	\caption{Reconstructed image quality vs. file size for various combinations of compression ratio and quantization step (Barbara, 2D-WHT, GAP-TV). Each curve corresponds to a particular compression ratio, \ie, CSr = \{0.02, 0.03, 0.04, 0.05, 0.07, 0.10, 0.15, 0.20\} and the points (markers) on each curve represent different quantization steps. From left to right, the quantization step decreases from 12 to 1 on each marker.} 
	\vspace{-0.4cm}
	\label{Fig:Joint_control}
\end{figure}


\subsection{Analysis of Joint Quality Control \label{Sec:result_jointcontrol}}
The effect of quantization on performance can be seen in Fig.~\ref{Fig:Joint_control}. In general, as the quantization step is decreased, the compressed file size and the reconstructed image quality are increased. However, the rate of increase is not constant. For each compression ratio there seems to be three distinct regions for the quantization step. In the lower quality/data rate region, decreasing the quantization step results in a significant gain in quality with little increase in the compressed file size; in the higher quality/data rate region the opposite is true: decreasing the quantization step increases the file size significantly with hardly any quality gain; and in the narrow intermediate region between those two we get moderate increase in quality and file size as the quantization step is decreased. The intermediate region occurs at higher quantization step for lower compression ratios. 
The highest possible reconstruction quality of Eq.~\eqref{Eq:RSN} is the maximum of all the curves depicted in Fig.~\ref{Fig:Joint_control} (as well as any potential curve of intermediate compression ratio). It is clear from the figure that at each bit rate $b$, the maximum is achieved in a different curve and $b$ is at the center of the intermediate region of that curve.

As mentioned earlier in Section~\ref{Sec:QualityCon}, after extensive analysis of the eight images shown in Fig.~\ref{fig:SSIM_8img_3algo}, we determined empirically that the maximum is achieved when the compression ratio and quantization step satisfy~\eqref{Eq:RSN}. While the optimal value for the constant $C$ in the right hand side varies from picture to picture, we found that $C=2.0$ gave good results for all images and we used it in our experiments on 200 images from the BSDS dataset~\cite{MartinFTM01}.


\begin{figure*}[tbp!]
	\centering
	\includegraphics[width=1\textwidth]{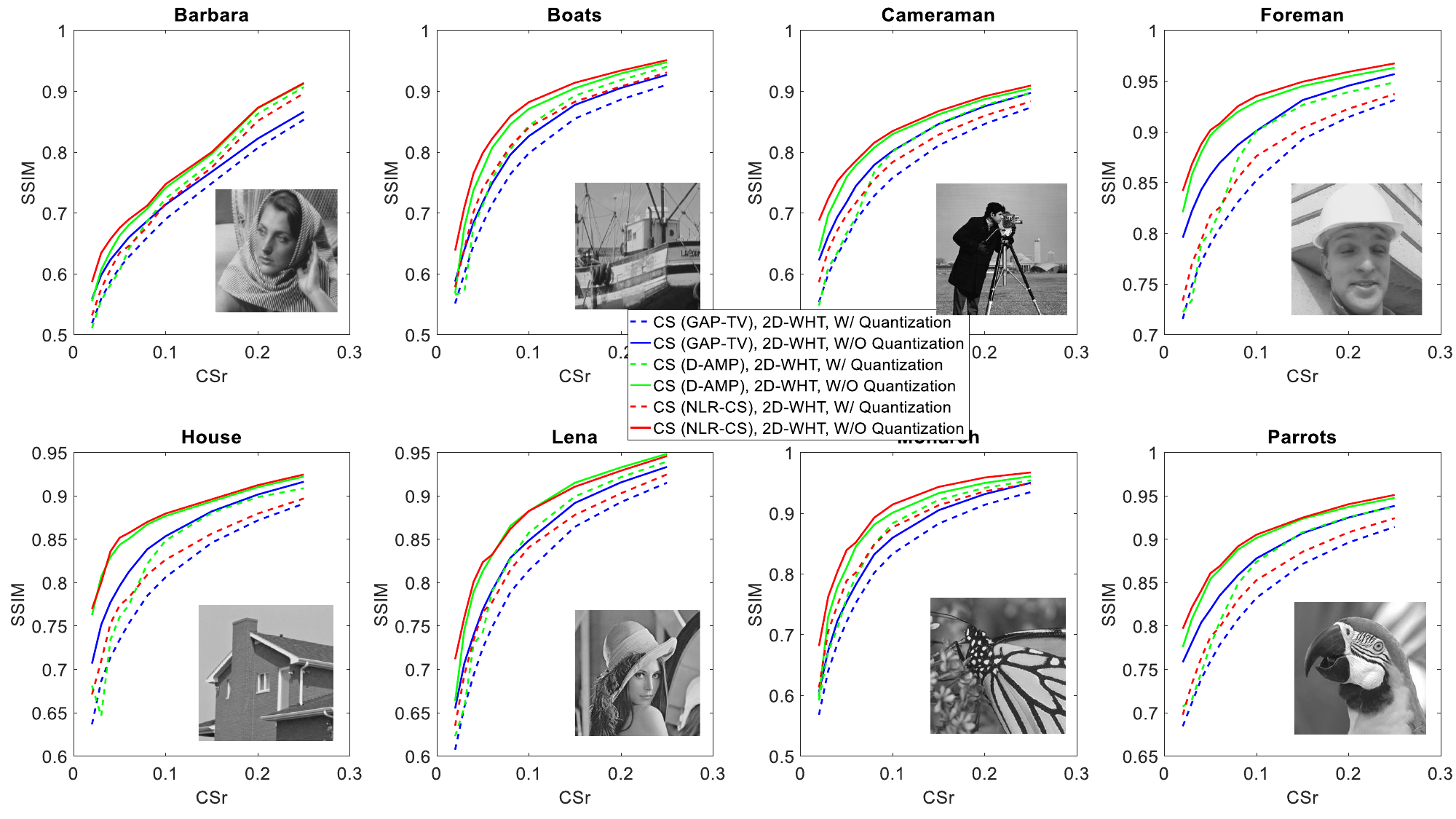}
	\vspace{-5mm}
	\caption{SSIM vs. compression ratio (CSr), comparing the reconstructed image with quantization (dash lines) and without quantization (solid lines) for different reconstruction algorithms. Please refer to~\cite{CSvsJPEG_webpage} for a larger plot.}
	\label{fig:SSIM_quan_err}
\end{figure*}
\subsection{Effect of the Reconstruction Algorithm \label{Sec:results_algo}}
We found that the more modern reconstruction algorithms, NLR-CS and D-AMP, give a better performance than GAP-TV (Fig.~\ref{fig:SSIM_8img_3algo}). In most cases NLR-CS is a little better than D-AMP, especially at low-quality/low bit rate settings. 
We note however, that both NLR-CS and D-AMP require careful manual parameter tuning and data normalization in order to perform well, so it is possible that the results would be somewhat different with better tuning.

The performance gains of D-AMP, and NLR-CS come at the cost of a much higher complexity. The reconstruction times of GAP-TV, D-AMP and NLR-CS  were 0.6, 35 and 147 seconds receptively, with the same compression rate on the same Intel i7 CPU with 24G memory. The algorithms were implemented in Matlab. {We expect that an optimized implementation in a compiled programming language would reduce the computation time by one or two orders of magnitude, and a significant further improvement can be achieved by using a GPU.} However, it appears that even with optimized implementation the last two algorithms  would run considerably slower than GAP-TV, because they are inherently more complex.
{Therefore, the choice of reconstruction algorithm can present a trade-off between quality and execution speed. Using different reconstruction algorithms, an implementation can offer a ``fast mode" for quick browsing and an ``accurate mode" for scrutinizing blown up images.}

\subsection{Effect of Quantization \label{Sec:quantization}}

{In order to get a better insight into the performance of our system, we isolate the component of the error which is due to quantization by running tests where the measurements vectors are reconstructed without quantization. Of course, in this case we cannot draw the usual performance curves of SSIM vs. compressed image file size, because without quantization (and entropy coding), there is no meaningful ``compressed image file''. Instead, we draw curves of quality, with and without quantization, vs. the compression ratio (CSr).} In Fig.~\ref{fig:SSIM_quan_err} we plot the SSIM of reconstructed images processed by CSbIC (including quantization), with that of images reconstructed directly from the unquantized measurements vectors. As expected,  the results without quantization are consistently better than the results with quantization. {In SSIM terms, the total decoding error is the difference between 1, which represents perfect reconstruction, and the SSIM value of the image decoded by the CSbIC. The error component due to quantization is the gap between curves with and without quantization}. This gap varies with the reconstruction algorithms and compression ratio, both in absolute terms and relative to the total decoding error. GAP-TV is not that sensitive to quantization error, with a SSIM decrease of 0.02--0.07 due to quantization. NLR-CS is the most sensitive, with SSIM decrease of 0.04--0.10. The behavior of D-AMP is not uniform: at compression ratios below 0.10 the SSIM decrease is similar to NLR-CS, but at higher compression ratios the SSIM difference decreases, down to 0.01--0.04 at a compression ratio of 0.25.  

\subsection{Comparison with JPEG and JPEG 2000 \label{Sec:JPEG_JPEG2000}}
CSbIC was tested with various sensing matrices and various reconstruction algorithms. As might be expected, the best combination depends on the particular image as well as on the quality or data rate operating point. However, in general the best performance was achieved with  2D-DCT+NLR-CS or 2D-WHT+D-AMP. In almost all test images CSbIC is decidedly better than JPEG at low bit rates (Fig.~\ref{fig:SSIM_8img_3algo}). A visual inspection shows (Fig.~\ref{fig:imges_cs_JPEG}) that at low bit rates JPEG images have blocking artifacts, while in CSbIC the artifacts are more subtle. 

In half of the images CS compression is better than JPEG for all quality levels. In the rest of the images JPEG is better for the higher quality cases, with the crossover happening at SSIM of 0.7--0.9 (0.8--0.9 for two out of the four images).

At this stage, the performance CSbIC is inferior to that of JPEG2000, but at the low quality regions, below about 0.7 SSIM, their performance is quite similar (Fig.~\ref{fig:SSIM_8img_3algo}). 

\section{Discussion \label{Sec:dis}}
Both JPEG and CS leverage the inherent compressibility of natural images, but in different ways: In JPEG this is done at the encoder --- the 2D DCT decomposition on $8\times 8$ pixel blocks essentially represents the signal according to a sparsity basis; in classical CS the compressibility assumption comes into play only at the reconstruction. JPEG also exploits another property of natural images, namely that the high magnitude coefficients are the low frequency ones, therefore JPEG allocates more bits to their accurate representation. This domain-specific insight was built into CSbIC by using the deterministic sensing matrices described in Sec.~\ref{Sec:SM_Mea}, which improved performance significantly in comparison to the SRMs of classical CS.

JPEG achieves lower bit rates solely through quantization, while CSbIC does it by quantization, and in addition, by discarding most of the transform coefficients during the measurements selection. This is possible because CS reconstruction recovers the missing coefficients by invoking various structural assumptions about the original image. This gives CSbIC a significant advantage at low bit rates, where excessive quantization results in poor performance of JPEG. Another possible explanation for that is that JPEG processes $8\times 8$ pixel blocks nearly independently. At low bit rates, only a small number of DCT coefficients are adequately represented in each JPEG block, which may be too little for blocks with much detail and causing the blocking artifacts which were observed. In contrast, all CS measurements contribute equally to the description of the whole image.

CSbIC yields better performance with certain types of sensing matrices and reconstruction algorithms. However, the less successful candidates also performed decently, and they may be the preferred choice in some situations. 
Binary-valued sensing matrices may be preferred for capturing measurements in the analog domain, and faster reconstruction may be useful when speed is important, for example, for quick image browsing.

JPEG is the culmination of many years of research into the properties of natural images. Wherever we could we leveraged this body of knowledge in CSbIC. However, much improvement can still be made, both in the quantization and entropy coding aspects of the system and in its CS aspects. The latter includes the sensing matrix design, the reconstruction algorithm, and the derivation of theoretical guarantees for successful reconstruction. The following are a few examples of such potential improvement in the area of reconstruction.

Recently, deep convolutional neural networks have been used, with good results, to solve the inverse problem (\ie, the reconstruction) in CS~\cite{Kulkarni2016CVPR,Yuan18OE,Ma_2019_ICCV,Miao_2019_ICCV,Lucas18SPM}.
This approach could dramatically reduce the reconstruction computation time, and potentially lead to improved performance, especially in domain-specific applications where the specifics of the domain may be difficult to express analytically. Such networks need to be trained on a very large image database. The quality of the reconstruction depends on the size and variability of the content of the database, which makes it difficult to make a fair comparison between the neural nets methods and the analytical reconstruction methods discussed in this paper. However, this is one of the most promising directions for improving the system.
As an additional point, the snapshot compressive video imaging systems~\cite{Patrick13OE,Yuan14CVPR} along with the recently developed algorithms~\cite{Ma_2019_ICCV,Liu18TPAMI,Yang14GMM,Yang14GMMonline} and the theoretical guarantees derived in~\cite{Jalali19TIT} demonstrate a promising way for video coding based on CS. This will be one of our future directions and can also be used for hyperspectral image~\cite{Yuan15JSTSP,Cao16SPM} compression.

\section*{Acknowledgments}
The authors would like to thank Dr. Lawrence O'Gorman at Bell Labs for help with the English editing and Paul Wilford at Bell Labs for helpful discussions.     

\bibliographystyle{IEEEtran}

\begin{thebibliography}{10}
	\providecommand{\url}[1]{#1}
	\csname url@samestyle\endcsname
	\providecommand{\newblock}{\relax}
	\providecommand{\bibinfo}[2]{#2}
	\providecommand{\BIBentrySTDinterwordspacing}{\spaceskip=0pt\relax}
	\providecommand{\BIBentryALTinterwordstretchfactor}{4}
	\providecommand{\BIBentryALTinterwordspacing}{\spaceskip=\fontdimen2\font plus
		\BIBentryALTinterwordstretchfactor\fontdimen3\font minus
		\fontdimen4\font\relax}
	\providecommand{\BIBforeignlanguage}[2]{{%
			\expandafter\ifx\csname l@#1\endcsname\relax
			\typeout{** WARNING: IEEEtran.bst: No hyphenation pattern has been}%
			\typeout{** loaded for the language `#1'. Using the pattern for}%
			\typeout{** the default language instead.}%
			\else
			\language=\csname l@#1\endcsname
			\fi
			#2}}
	\providecommand{\BIBdecl}{\relax}
	\BIBdecl
	
	\bibitem{CSvsJPEG_webpage}
	``{CS vs. JPEG},'' \url{https://sites.google.com/site/eiexyuan/csvsjpeg}.
	
	\bibitem{Candes06ITT}
	E.~J. Cand\`{e}s, J.~Romberg, and T.~Tao, ``Robust uncertainty principles:
	Exact signal reconstruction from highly incomplete frequency information,''
	\emph{IEEE Transactions on Information Theory}, vol.~52, no.~2, pp. 489--509,
	February 2006.
	
	\bibitem{Donoho06ITT}
	D.~L. Donoho, ``Compressed sensing,'' \emph{IEEE Transactions on Information
		Theory}, vol.~52, no.~4, pp. 1289--1306, April 2006.
	
	\bibitem{Elad10_sparse}
	M.~Elad, \emph{Sparse and Redundant Representations: From Theory to
		Applications in Signal and Image Processing}.\hskip 1em plus 0.5em minus
	0.4em\relax Springer, 2010.
	
	\bibitem{Dong14TIP}
	W.~Dong, G.~Shi, X.~Li, Y.~Ma, and F.~Huang, ``Compressive sensing via nonlocal
	low-rank regularization,'' \emph{IEEE Transactions on Image Processing},
	vol.~23, no.~8, pp. 3618--3632, 2014.
	
	\bibitem{Mertzler14Denoising}
	C.~A. Metzler, A.~Maleki, and R.~G. Baraniuk, ``From denoising to compressed
	sensing,'' \emph{IEEE Transactions on Information Theory}, vol.~62, no.~9,
	pp. 5117--5144, Sept 2016.
	
	\bibitem{Yuan16TMM_privacyCS}
	X.~{Yuan}, X.~{Wang}, C.~{Wang}, J.~{Weng}, and K.~{Ren}, ``Enabling secure and
	fast indexing for privacy-assured healthcare monitoring via compressive
	sensing,'' \emph{IEEE Transactions on Multimedia}, vol.~18, no.~10, pp.
	2002--2014, Oct 2016.
	
	\bibitem{Song17TMM_CScloud}
	X.~{Song}, X.~{Peng}, J.~{Xu}, G.~{Shi}, and F.~{Wu}, ``Distributed compressive
	sensing for cloud-based wireless image transmission,'' \emph{IEEE
		Transactions on Multimedia}, vol.~19, no.~6, pp. 1351--1364, June 2017.
	
	\bibitem{Chen18TMM_BCS}
	Z.~{Chen}, X.~{Hou}, X.~{Qian}, and C.~{Gong}, ``Efficient and robust image
	coding and transmission based on scrambled block compressive sensing,''
	\emph{IEEE Transactions on Multimedia}, vol.~20, no.~7, pp. 1610--1621, July
	2018.
	
	\bibitem{Liu16TMM_PCA}
	Y.~{Liu} and D.~A. {Pados}, ``Compressed-sensed-domain l1-pca video
	surveillance,'' \emph{IEEE Transactions on Multimedia}, vol.~18, no.~3, pp.
	351--363, March 2016.
	
	\bibitem{Liu14TMM_BitVideo}
	H.~{Liu}, B.~{Song}, F.~{Tian}, and H.~{Qin}, ``Joint sampling rate and
	bit-depth optimization in compressive video sampling,'' \emph{IEEE
		Transactions on Multimedia}, vol.~16, no.~6, pp. 1549--1562, Oct 2014.
	
	\bibitem{Zhang16TMM_BiCS}
	L.~Y. {Zhang}, K.~{Wong}, Y.~{Zhang}, and J.~{Zhou}, ``Bi-level protected
	compressive sampling,'' \emph{IEEE Transactions on Multimedia}, vol.~18,
	no.~9, pp. 1720--1732, Sep. 2016.
	
	\bibitem{Deng12TMM}
	C.~{Deng}, W.~{Lin}, B.~{Lee}, and C.~T. {Lau}, ``Robust image coding based
	upon compressive sensing,'' \emph{IEEE Transactions on Multimedia}, vol.~14,
	no.~2, pp. 278--290, April 2012.
	
	\bibitem{Goyal08SPM}
	V.~K. Goyal, A.~K. Fletcher, and S.~Rangan, ``Compressive sampling and lossy
	compression,'' \emph{IEEE Signal Processing Magazine}, vol.~25, no.~2, pp.
	48--96, 2008.
	
	\bibitem{Laska12_TSP_bitchange}
	J.~N. Laska and R.~G. Baraniuk, ``Regime change: Bit-depth versus
	measurement-rate in compressive sensing,'' \emph{IEEE Transactions on Signal
		Processing}, vol.~60, no.~7, pp. 3496--3505, July 2012.
	
	\bibitem{LASKA11_ACHA}
	J.~N. Laska, P.~T. Boufounos, M.~A. Davenport, and R.~G. Baraniuk, ``Democracy
	in action: Quantization, saturation, and compressive sensing,'' \emph{Applied
		and Computational Harmonic Analysis}, vol.~31, no.~3, pp. 429 -- 443, 2011.
	
	\bibitem{Zymnis10SPL}
	A.~Zymnis, S.~Boyd, and E.~Candes, ``Compressed sensing with quantized
	measurements,'' \emph{IEEE Signal Processing Letters}, vol.~17, no.~2, pp.
	149--152, Feb 2010.
	
	\bibitem{Laska11_TSP_1bit}
	J.~N. Laska, Z.~Wen, W.~Yin, and R.~G. Baraniuk, ``Trust, but verify: Fast and
	accurate signal recovery from 1-bit compressive measurements,'' \emph{IEEE
		Transactions on Signal Processing}, vol.~59, no.~11, pp. 5289--5301, Nov.
	2011.
	
	\bibitem{Dai09_ISIT}
	W.~{Dai}, H.~V. {Pham}, and O.~{Milenkovic}, ``A comparative study of quantized
	compressive sensing schemes,'' in \emph{2009 IEEE International Symposium on
		Information Theory}, June 2009, pp. 11--15.
	
	\bibitem{Dai09_ITW}
	------, ``Distortion-rate functions for quantized compressive sensing,'' in
	\emph{2009 IEEE Information Theory Workshop on Networking and Information
		Theory}, June 2009, pp. 171--175.
	
	\bibitem{Baig10_ICT}
	Y.~{Baig}, E.~M. {Lai}, and J.~P. {Lewis}, ``Quantization effects on compressed
	sensing video,'' in \emph{2010 17th International Conference on
		Telecommunications}, April 2010, pp. 935--940.
	
	\bibitem{Venkatraman09_ICASSP}
	D.~Venkatraman and A.~Makur, ``A compressive sensing approach to object-based
	surveillance video coding,'' in \emph{2009 IEEE International Conference on
		Acoustics, Speech and Signal Processing}, April 2009, pp. 3513--3516.
	
	\bibitem{Cui18MM}
	W.~Cui, F.~Jiang, X.~Gao, S.~Zhang, and D.~Zhao, ``An efficient deep quantized
	compressed sensing coding framework of natural images,'' in \emph{Proceedings
		of the 26th ACM International Conference on Multimedia}, 2018, pp.
	1777--1785.
	
	\bibitem{JPEG1994}
	{Information Technology -- Digital Compression And Coding Of Continuous Tone
		Still Images -- Requirements And Guidelines}, ``{CCITT Recommendation
		T.81},'' 1992.
	
	\bibitem{JPEG2000}
	{Information technology -- JPEG 2000 image coding system: Core coding system},
	``{ITU-T Recommendation T.800},'' 2002.
	
	\bibitem{Hannuksela2015TheHE}
	M.~M. Hannuksela, J.~Lainema, and V.~K.~M. Vadakital, ``The high efficiency
	image file format standard [standards in a nutshell],'' \emph{IEEE Signal
		Processing Magazine}, vol.~32, pp. 150--156, 2015.
	
	\bibitem{Lainema2016HEVCSI}
	J.~Lainema, M.~M. Hannuksela, V.~K.~M. Vadakital, and E.~B. Aksu, ``Hevc still
	image coding and high efficiency image file format,'' \emph{2016 IEEE
		International Conference on Image Processing (ICIP)}, pp. 71--75, 2016.
	
	\bibitem{Ginesu2012ObjectiveAO}
	G.~Ginesu, M.~Pintus, and D.~D. Giusto, ``Objective assessment of the webp
	image coding algorithm,'' \emph{Sig. Proc.: Image Comm.}, vol.~27, pp.
	867--874, 2012.
	
	\bibitem{Xu14TCSVT}
	M.~{Xu}, S.~{Li}, J.~{Lu}, and W.~{Zhu}, ``Compressibility constrained sparse
	representation with learnt dictionary for low bit-rate image compression,''
	\emph{IEEE Transactions on Circuits and Systems for Video Technology},
	vol.~24, no.~10, pp. 1743--1757, Oct 2014.
	
	\bibitem{Zhang17DCC}
	X.~{Zhang}, S.~{Ma}, Z.~{Lin}, J.~{Zhang}, S.~{Wang}, and W.~{Gao}, ``Globally
	variance-constrained sparse representation for rate-distortion optimized
	image representation,'' in \emph{Data Compression Conference (DCC)}, April
	2017, pp. 380--389.
	
	\bibitem{Zhang18TIP}
	X.~{Zhang}, J.~{Sun}, S.~{Ma}, Z.~{Lin}, J.~{Zhang}, S.~{Wang}, and W.~{Gao},
	``Globally variance-constrained sparse representation and its application in
	image set coding,'' \emph{IEEE Transactions on Image Processing}, vol.~27,
	no.~8, pp. 3753--3765, August 2018.
	
	\bibitem{Duarte08SPM}
	M.~F. Duarte, M.~A. Davenport, D.~Takhar, J.~N. Laska, T.~Sun, K.~F. Kelly, and
	R.~G. Baraniuk, ``Single-pixel imaging via compressive sampling,'' \emph{IEEE
		Signal Processing Magazine}, vol.~25, no.~2, pp. 83--91, 2008.
	
	\bibitem{Huang13ICIP}
	G.~{Huang}, H.~{Jiang}, K.~{Matthews}, and P.~{Wilford}, ``Lensless imaging by
	compressive sensing,'' in \emph{2013 IEEE International Conference on Image
		Processing}, Sep. 2013, pp. 2101--2105.
	
	\bibitem{Yuan18OE}
	X.~Yuan and Y.~Pu, ``Parallel lensless compressive imaging via deep
	convolutional neural networks,'' \emph{Optics Express}, vol.~26, no.~2, pp.
	1962--1977, Jan 2018.
	
	\bibitem{Wang04imagequality}
	Z.~Wang, A.~C. Bovik, H.~R. Sheikh, and E.~P. Simoncelli, ``Image quality
	assessment: From error visibility to structural similarity,'' \emph{IEEE
		Transactions on Image Processing}, vol.~13, no.~4, pp. 600--612, 2004.
	
	\bibitem{Romberg08SPM}
	J.~Romberg, ``Imaging via compressive sampling,'' \emph{IEEE Signal Processing
		Magazine}, vol.~25, no.~2, pp. 14--20, 2008.
	
	\bibitem{ahn2016compressive}
	J.-H. Ahn, ``Compressive sensing and recovery for binary images,'' \emph{IEEE
		Transactions on Image Processing}, vol.~25, no.~10, pp. 4796--4802, 2016.
	
	\bibitem{Yuan15Lensless}
	X.~Yuan, H.~Jiang, G.~Huang, and P.~Wilford, ``Lensless compressive imaging,''
	\emph{arXiv:1508.03498}, 2015.
	
	\bibitem{Yuan16SJ}
	------, ``{SLOPE}: Shrinkage of local overlapping patches estimator for
	lensless compressive imaging,'' \emph{IEEE Sensors Journal}, vol.~16, no.~22,
	pp. 8091--8102, November 2016.
	
	\bibitem{Pratt69_IEEE}
	W.~K. Pratt, J.~Kane, and H.~C. Andrews, ``Hadamard transform image coding,''
	\emph{Proceedings of the IEEE}, vol.~57, no.~1, pp. 58--68, Jan 1969.
	
	\bibitem{Duarte12_KroCS}
	M.~F. Duarte and R.~G. Baraniuk, ``Kronecker compressive sensing,'' \emph{IEEE
		Transactions on Image Processing}, vol.~21, no.~2, pp. 494--504, Feb 2012.
	
	\bibitem{Fino76_TC_WHT}
	B.~J. Fino and V.~R. Algazi, ``Unified matrix treatment of the fast
	walsh-hadamard transform,'' \emph{IEEE Transactions on Computers}, vol. C-25,
	no.~11, pp. 1142--1146, Nov 1976.
	
	\bibitem{cs_Candes06randomProj}
	E.~J. Candes and T.~Tao, ``Near-optimal signal recovery from random
	projections: Universal encoding strategies?'' \emph{IEEE Transactions on
		Information Theory}, vol.~52, no.~12, pp. 5406--5425, Dec. 2006.
	
	\bibitem{Candes05_LinearP}
	------, ``Decoding by linear programming,'' \emph{IEEE Transactions on
		Information Theory}, vol.~51, no.~12, pp. 4203--4215, Dec 2005.
	
	\bibitem{Haimi-CohenL16_SP}
	R.~Haimi-Cohen and Y.~M. Lai, ``Compressive measurements generated by
	structurally random matrices: Asymptotic normality and quantization.''
	\emph{Signal Processing}, vol. 120, pp. 71--87, 2016.
	
	\bibitem{Do12_SRM}
	T.~T. Do, L.~Gan, N.~H. Nguyen, and T.~D. Tran, ``Fast and efficient
	compressive sensing using structurally random matrices,'' \emph{IEEE
		Transactions on Signal Processing}, vol.~60, no.~1, pp. 139--154, Jan 2012.
	
	\bibitem{Do08_ICASSP}
	T.~T. Do, T.~D. Tran, and L.~Gan, ``Fast compressive sampling with structurally
	random matrices,'' in \emph{2008 IEEE International Conference on Acoustics,
		Speech and Signal Processing}, March 2008, pp. 3369--3372.
	
	\bibitem{Gersho1991}
	A.~Gersho and R.~M. Gray, \emph{Vector Quantization and Signal
		Compression}.\hskip 1em plus 0.5em minus 0.4em\relax Norwell, MA, USA: Kluwer
	Academic Publishers, 1991.
	
	\bibitem{Kipnis17ISIT}
	A.~Kipnis, G.~Reeves, Y.~C. Eldar, and A.~J. Goldsmith, ``Compressed sensing
	under optimal quantization,'' in \emph{2017 IEEE International Symposium on
		Information Theory (ISIT)}, June 2017, pp. 2148--2152.
	
	\bibitem{Kipnis16TIT}
	A.~Kipnis, A.~J. Goldsmith, Y.~C. Eldar, and T.~Weissman, ``Distortion rate
	function of sub-nyquist sampled gaussian sources,'' \emph{IEEE Transactions
		on Information Theory}, vol.~62, no.~1, pp. 401--429, Jan 2016.
	
	\bibitem{Kipnis16ADC}
	A.~J.~G. Alon~Kipnis, Yonina C.~Eldar, ``Fundamental distortion limits of
	analog-to-digital compression,'' arXiv:1601.06421, Tech. Rep., 2016.
	
	\bibitem{Candes11RIPless}
	E.~J. Candes and Y.~Plan, ``A probabilistic and ripless theory of compressed
	sensing,'' \emph{IEEE Transactions on Information Theory}, vol.~57, no.~11,
	pp. 7235--7254, Nov 2011.
	
	\bibitem{MartinFTM01}
	D.~Martin, C.~Fowlkes, D.~Tal, and J.~Malik, ``A database of human segmented
	natural images and its application to evaluating segmentation algorithms and
	measuring ecological statistics,'' in \emph{Proc. 8th Int'l Conf. Computer
		Vision}, vol.~2, July 2001, pp. 416--423.
	
	\bibitem{Langdon79arithmeticcoding}
	G.~G. Langdon, ``Arithmetic coding,'' \emph{IBM J. Res. Develop}, vol.~23, pp.
	149--162, 1979.
	
	\bibitem{Aharon06TSP}
	M.~Aharon, M.~Elad, and A.~Bruckstein, ``{K-SVD}: An algorithm for designing
	overcomplete dictionaries for sparse representation,'' \emph{IEEE
		Transactions on Signal Processing}, vol.~54, no.~11, pp. 4311--4322, 2006.
	
	\bibitem{Yuan15GMM}
	X.~Yuan, H.~Jiang, G.~Huang, and P.~Wilford, ``Compressive sensing via low-rank
	{Gaussian} mixture models,'' \emph{arXiv:1508.06901}, 2015.
	
	\bibitem{Zhang18CVPR}
	X.~Zhang, X.~Yuan, and L.~Carin, ``Nonlocal low-rank tensor factor analysis for
	image restoration,'' in \emph{IEEE Conference on Computer Vision and Pattern
		Recognition (CVPR)}, 2018, pp. 3318--3325.
	
	\bibitem{Zha2020RRC_TIP}
	Z.~{Zha}, X.~{Yuan}, B.~{Wen}, J.~{Zhou}, J.~{Zhang}, and C.~{Zhu}, ``From rank
	estimation to rank approximation: Rank residual constraint for image
	restoration,'' \emph{IEEE Transactions on Image Processing}, pp. 1--1, 2019.
	
	\bibitem{Yuan18SP}
	X.~Yuan, ``Adaptive step-size iterative algorithm for sparse signal recovery,''
	\emph{Signal Processing}, vol. 152, pp. 273--285, 2018.
	
	\bibitem{Dabov07BM3D}
	K.~Dabov, A.~Foi, V.~Katkovnik, and K.~Egiazarian, ``Image denoising by sparse
	3d transform-domain collaborative filtering,'' \emph{IEEE Transactions on
		Image Processing}, vol.~16, no.~8, pp. 2080--2095, August 2007.
	
	\bibitem{Donoho09AMP}
	D.~L. Donoho, A.~Maleki, and A.~Montanari, ``Message-passing algorithms for
	compressed sensing,'' \emph{Proceedings of the National Academy of Sciences},
	vol. 106, no.~45, pp. 18\,914--18\,919, 2009.
	
	\bibitem{Boyd11ADMM}
	S.~Boyd, N.~Parikh, E.~Chu, B.~Peleato, and J.~Eckstein, ``Distributed
	optimization and statistical learning via the alternating direction method of
	multipliers,'' \emph{Found. Trends Mach. Learn.}, vol.~3, no.~1, pp. 1--122,
	January 2011.
	
	\bibitem{Zhang14TIP}
	J.~Zhang, D.~Zhao, and W.~Gao, ``Group-based sparse representation for image
	restoration,'' \emph{IEEE Transactions on Image Processing}, vol.~23, no.~8,
	pp. 3336--3351, August 2014.
	
	\bibitem{Shi17ICME}
	W.~{Shi}, F.~{Jiang}, S.~{Zhang}, and D.~{Zhao}, ``Deep networks for compressed
	image sensing,'' in \emph{2017 IEEE International Conference on Multimedia
		and Expo (ICME)}, July 2017, pp. 877--882.
	
	\bibitem{Yuan16ICIP_GAP}
	X.~Yuan, ``Generalized alternating projection based total variation
	minimization for compressive sensing,'' in \emph{IEEE International
		Conference on Image Processing}, Sept 2016, pp. 2539--2543.
	
	\bibitem{Kulkarni2016CVPR}
	K.~{Kulkarni}, S.~{Lohit}, P.~{Turaga}, R.~{Kerviche}, and A.~{Ashok},
	``Reconnet: Non-iterative reconstruction of images from compressively sensed
	measurements,'' in \emph{2016 IEEE Conference on Computer Vision and Pattern
		Recognition (CVPR)}, June 2016, pp. 449--458.
	
	\bibitem{Ma_2019_ICCV}
	J.~Ma, X.-Y. Liu, Z.~Shou, and X.~Yuan, ``Deep tensor admm-net for snapshot
	compressive imaging,'' in \emph{The IEEE International Conference on Computer
		Vision (ICCV)}, October 2019.
	
	\bibitem{Miao_2019_ICCV}
	X.~Miao, X.~Yuan, Y.~Pu, and V.~Athitsos, ``$\lambda$-net: Reconstruct
	hyperspectral images from a snapshot measurement,'' in \emph{The IEEE
		International Conference on Computer Vision (ICCV)}, October 2019.
	
	\bibitem{Lucas18SPM}
	A.~{Lucas}, M.~{Iliadis}, R.~{Molina}, and A.~K. {Katsaggelos}, ``Using deep
	neural networks for inverse problems in imaging: Beyond analytical methods,''
	\emph{IEEE Signal Processing Magazine}, vol.~35, no.~1, pp. 20--36, January
	2018.
	
	\bibitem{Patrick13OE}
	P.~Llull, X.~Liao, X.~Yuan, J.~Yang, D.~Kittle, L.~Carin, G.~Sapiro, and D.~J.
	Brady, ``Coded aperture compressive temporal imaging,'' \emph{Optics
		Express}, pp. 10\,526--10\,545, 2013.
	
	\bibitem{Yuan14CVPR}
	X.~Yuan, P.~Llull, X.~Liao, J.~Yang, G.~Sapiro, D.~J. Brady, and L.~Carin,
	``Low-cost compressive sensing for color video and depth,'' in \emph{IEEE
		Conference on Computer Vision and Pattern Recognition (CVPR)}, 2014.
	
	\bibitem{Liu18TPAMI}
	Y.~{Liu}, X.~{Yuan}, J.~{Suo}, D.~J. {Brady}, and Q.~{Dai}, ``Rank minimization
	for snapshot compressive imaging,'' \emph{IEEE Transactions on Pattern
		Analysis and Machine Intelligence}, vol.~41, no.~12, pp. 2990--3006, Dec
	2019.
	
	\bibitem{Yang14GMM}
	J.~Yang, X.~Yuan, X.~Liao, P.~Llull, G.~Sapiro, D.~J. Brady, and L.~Carin,
	``Video compressive sensing using {G}aussian mixture models,'' \emph{IEEE
		Transaction on Image Processing}, vol.~23, no.~11, pp. 4863--4878, November
	2014.
	
	\bibitem{Yang14GMMonline}
	J.~Yang, X.~Liao, X.~Yuan, P.~Llull, D.~J. Brady, G.~Sapiro, and L.~Carin,
	``Compressive sensing by learning a {G}aussian mixture model from
	measurements,'' \emph{IEEE Transaction on Image Processing}, vol.~24, no.~1,
	pp. 106--119, January 2015.
	
	\bibitem{Jalali19TIT}
	S.~{Jalali} and X.~{Yuan}, ``Snapshot compressed sensing: Performance bounds
	and algorithms,'' \emph{IEEE Transactions on Information Theory}, vol.~65,
	no.~12, pp. 8005--8024, Dec 2019.
	
	\bibitem{Yuan15JSTSP}
	X.~Yuan, T.-H. Tsai, R.~Zhu, P.~Llull, D.~J. Brady, and L.~Carin, ``Compressive
	hyperspectral imaging with side information,'' \emph{IEEE Journal of Selected
		Topics in Signal Processing}, vol.~9, no.~6, pp. 964–--976, 2015.
	
	\bibitem{Cao16SPM}
	X.~Cao, T.~Yue, X.~Lin, S.~Lin, X.~Yuan, Q.~Dai, L.~Carin, and D.~J. Brady,
	``Computational snapshot multispectral cameras: Toward dynamic capture of the
	spectral world,'' \emph{IEEE Signal Processing Magazine}, vol.~33, no.~5, pp.
	95--108, Sept 2016.
	
\end{thebibliography}

\end{document}